\documentclass[11pt]{elsarticle}
\usepackage{placeins}
\usepackage{lineno,hyperref,color}
\usepackage{xcolor}
\usepackage{mwe}

\modulolinenumbers[5]
\definecolor{LightCyan}{rgb}{0.5,0.3,0.1}
\definecolor{Gray}{rgb}{0.7,0.8,1}
\journal{Pattern Recognition}

\usepackage{breqn}
\usepackage{amsmath,amssymb}
\usepackage{ulem}
 \usepackage{float}
\usepackage{multirow}

\usepackage{pifont}
\newcommand{\cmark}{\text{\ding{51}}}

\begin{document}

\begin{frontmatter}



\title{Enhance to Read Better: A Multi-Task Adversarial Network for Handwritten Document Image Enhancement}


\author[mymainaddress3,mymainaddress1]{Sana Khamekhem Jemni\corref{equalcontrub}}
\author[mymainaddress2]{Mohamed Ali Souibgui\corref{equalcontrub}}
\author[mymainaddress3,mymainaddress4]{Yousri Kessentini}
\author[mymainaddress2]{Alicia Fornés}

\address[mymainaddress1]{MIR@CL: Multimedia, InfoRmation systems and Advanced Computing Laboratory}
\address[mymainaddress2]{Computer Vision Center, Computer Science Department, Universitat Autònoma de Barcelona, Spain}
\address[mymainaddress3]{Digital Research Center of Sfax, B.P. 275, Sakiet Ezzit, 3021 Sfax, Tunisia}
\address[mymainaddress4]{SM@RTS : Laboratory of Signals, systeMs, aRtificial Intelligence and neTworkS}

\cortext[equalcontrub]{Those authors were equally contributed to this paper}


\begin{abstract}

Handwritten  document images can be highly affected by degradation for different reasons: Paper ageing, daily-life scenarios (wrinkles, dust, etc.), bad scanning process and so on. These artifacts raise many readability issues for current Handwritten Text Recognition (HTR) algorithms and severely devalue their efficiency. In this paper, we propose an end to end architecture based on Generative Adversarial Networks (GANs) to recover the degraded documents into a $clean$ and $readable$ form. Unlike the most well-known document binarization methods, which try to improve the visual quality of the degraded document, the proposed architecture integrates a handwritten text recognizer that promotes the generated document image to be more readable. To the best of our knowledge,  this is the first work to use the text information while binarizing handwritten documents. Extensive experiments conducted on degraded Arabic and Latin handwritten documents demonstrate the usefulness of integrating the recognizer within the GAN architecture, which improves both the visual quality and the readability of the degraded document images. Moreover, we outperform the state of the art in H-DIBCO challenges, after fine tuning our pre-trained model with synthetically degraded Latin handwritten images, on this task.    

\end{abstract}

\begin{keyword}
Handwritten Document Image Binarization \sep Document Enhancement \sep Handwriting Text Recognition \sep Generative Adversarial Networks  \sep Recurrent Neural Networks 
\end{keyword}

\end{frontmatter}


\section{Introduction}

Handwritten document analysis is an active and important field in computer vision and pattern recognition community. With the recent developments in machine learning \cite{goodfellow2016deep}, processing handwritten documents is reaching a good accuracy, especially in the application of Handwritten Text Recognition (HTR). HTR is the crucial part towards automatically understanding a document, which facilitates the access to various automatic applications such as: information extraction, search, indexation, validation, etc. 

\begin{figure}[t]
\begin{center}
\begin{tabular}{ccc}
\includegraphics[width=0.28\columnwidth, height=2.5cm]{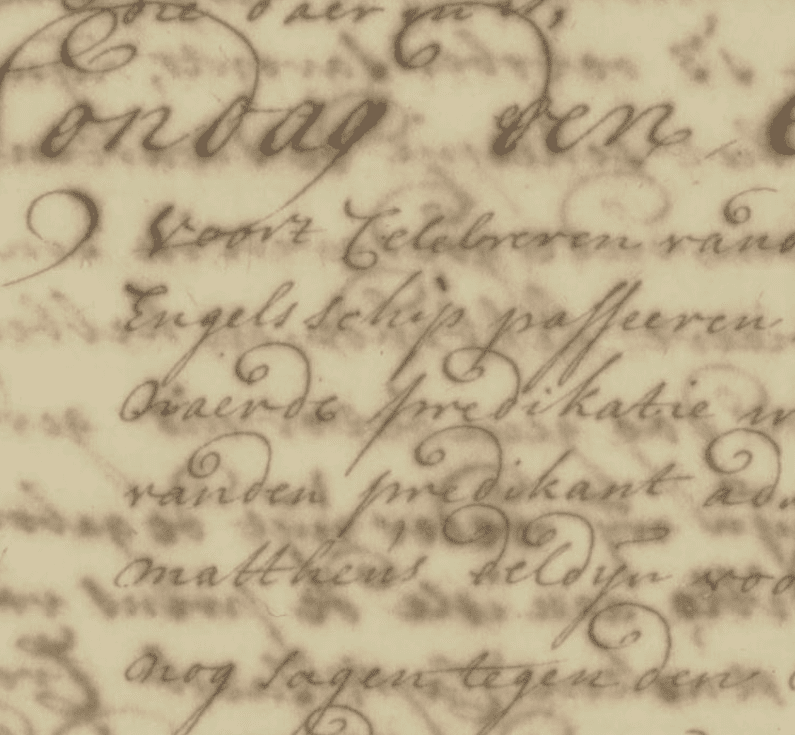}&
\includegraphics[width=0.28\columnwidth, height=2.5cm]{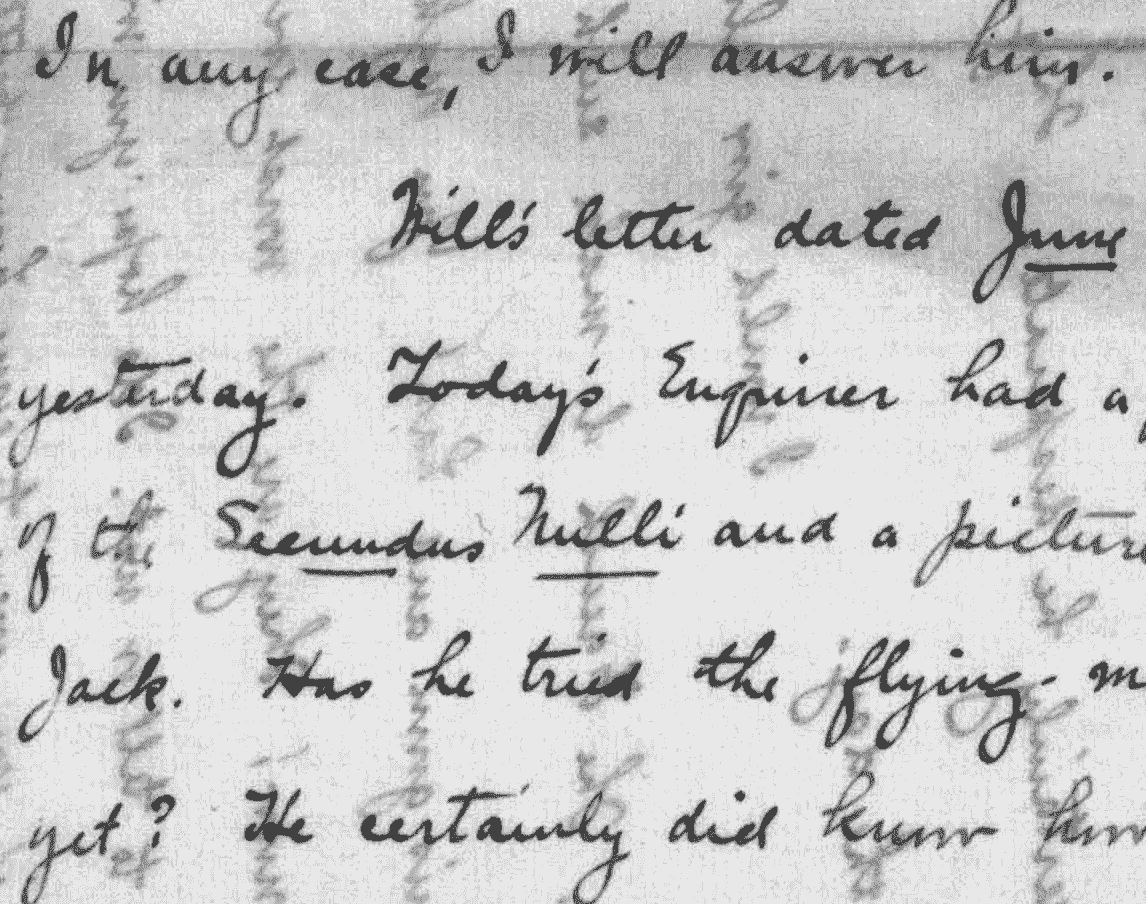}&
\includegraphics[width=0.28\columnwidth, height=2.5cm]{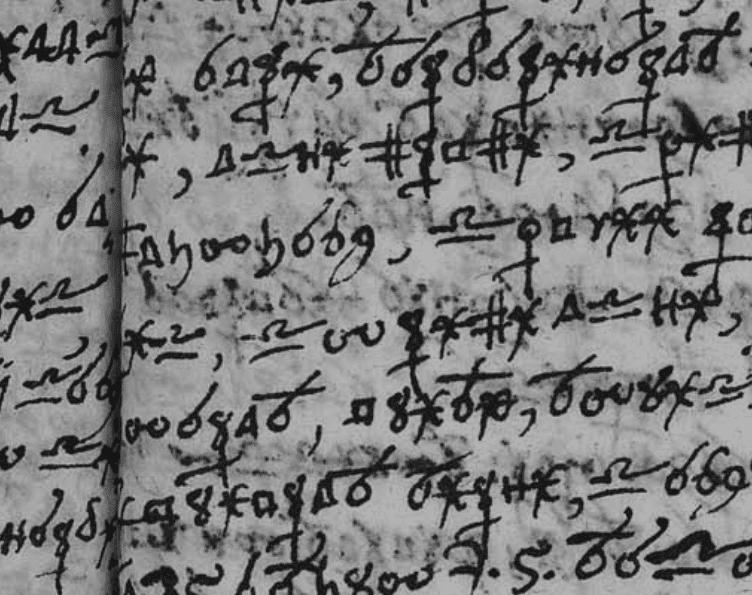} \\

\multicolumn{3}{l}{\includegraphics[width=0.91\columnwidth, height=2.5cm]{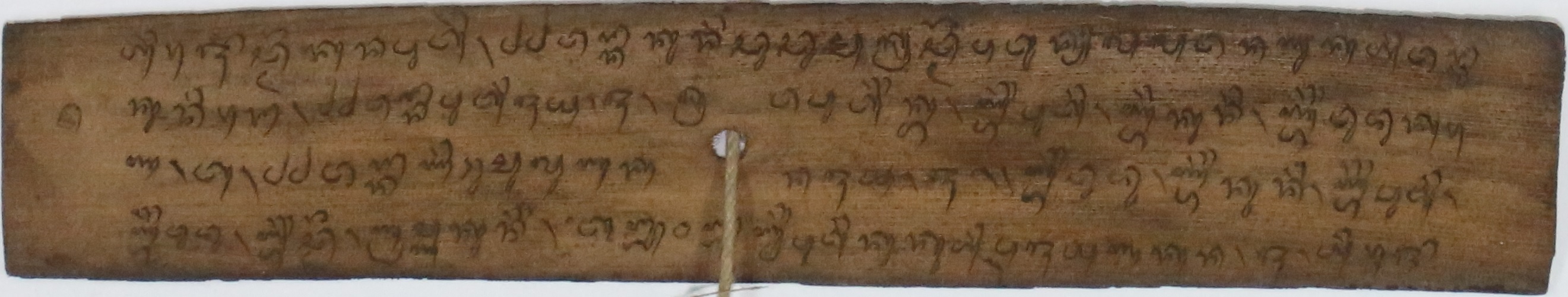}}

\end{tabular}
\caption{Examples of the degradation that can be appeared in handwritten text images.}
\label{fig:examples_used}
\end{center}
\end{figure}

One of the problems that HTR systems are facing  is the degradation of an inputted document, which significantly decreases the reading performance, reflecting on its utility. Indeed, many degradation scenarios can be attached to a handwritten document, especially the historical ones. Degradation includes  background noise, corrupted text, dust, wrinkles and historical effects just to name a few  related to the condition of the document itself \cite{Pratikakis2018icfhr}. The bad scanning process can also produce problems (shadows, blur, light distortion, angle, etc.) \cite{souibgui2020conditional,wang2019an}. Moreover, some documents contain watermarks or stamps inserted for security reasons, those can cover the text and  obstruct the HTR engine \cite{souibgui2020gan}. Some  degradation examples   are presented in Figure~\ref{fig:examples_used}, as it can be seen, cleaning the document before passing it to the HTR stage should be done. 


This cleaning task, called document enhancement, includes different recovering techniques, to reverse the degradation effect, for example: Binarization, dewarping, deblurring, watermark removal, etc.
Classic recovery techniques were integrating image processing algorithms to be used as a filter that separates the degradation from the text. However, those methods are failing in removing the high degradation. Also, their parameters are usually set depending on the quality state of the addressed document to produce an optimal result. Thus, a manual intervention is needed in some cases to adjust the parameters, which is quite costly.

Given  this, some modern document recovery techniques are appearing, using  machine learning tools.  Those  are  training deep learning models, mainly Convolutional Neural Networks (CNNs) and Generative Adversarial Networks (GANs), to learn the parameters for a direct mapping of any degraded document image into a clean binary version (without a restriction on degradation level) \cite{tamrin2021,kang2021complex,zhao2019document}. {Similar to those, we proposed in \cite{souibgui2020gan} a document enhancement model called DE-GAN. However, and despite the high accuracy that we achieved  in various  enhancement tasks (Binarization, cleaning, watermark removal, deblurring),} an important evaluation was not done. In fact, the goal of document enhancement is to provide a cleaner   version of the image which is highly beneficial for HTR engines. But, in the mentioned approaches (including ours), the evaluation was conducted using only   the visual similarity measurement between the  recovered image and the Ground Truth (GT) clean version with some metrics that depend on pixels values. 
Thus, a HTR evaluation (means, passing the images to a HTR engine and compare the recognized text with the GT) is missing, for a  better validation of the developed approaches. Also, these models are generally trained  using only the images, while ignoring the text. As result, a model can easily evolve to deteriorate the text while cleaning the degraded image.  

{Motivated by those challenges, we propose a new method  consists in an improved version of our previously developed DE-GAN, which was designed to recover the handwritten document to a clean version while ensuring its readability}. Our approach is a deep learning model based on GANs that learns its parameters not only from the handwritten images pairs (degraded + GT), but also from the associated  GT text. {For this aim, we propose to add a recognizer that is trained jointly in a GAN model to assess the readability of the recovered document image.}  Hence, the model shall learn the best mapping of the degraded image to be as clean as possible while keeping its text readable. To accomplish this, and since the used  datasets for document binarization does not (or rarely) contain the text information, we used two publicly available handwriting text images datasets (KHATT for Arabic script and IAM for Latin script) that are originally used for HTR to create degraded versions from the GT clean text lines images. 
The contribution of this paper can be summarized as follows: 
\begin{itemize}

    \item To the best of our knowledge, this is the first work that  integrates a recognition stage in a document binarization model. Thus, the degraded handwritten document will be recovered while maximising its readability, simultaneously.  This is done by combining the GAN and the Connectionist Temporal Classification (CTC) losses functions: We eliminate the noise while preserving the handwritten text strokes.
    \item { We demonstrate that training  the recognizer progressively  (on images ordered  from  the  degraded  domain  to the  clean  versions),  improves the recognition performance.}
    \item The proposed model is simple, and flexible to restore different forms of degradation, independently  of the document language. This was shown by the experiments conducted on two created datasets namely degraded-IAM (Latin script) and degraded-KHATT (Arabic script). 
    \item We achieved the SOTA performance in handwritten document  binarization according to H-DIBCO benchmarks.
\end{itemize}
 
    
   
    
 


The rest of this paper is organized as follows. Section~\ref{section:relatedwork} provides a review of prior works on document enhancement, in particular for document recovery and binarization. Then we present our proposed model in Section \ref{section:proposedapproach}. After that, experimental results and comparisons with recent methods will be described in Section \ref{section:experiments}. Finally, a conclusion with a future direction is given in Section \ref{section:conclusion}.

\section{Related Work}\label{section:relatedwork}

 This work aims to recover images that contain hard degradation by removing the background noise, while keeping its readability by HTR systems as accurate as possible. This application called document enhancement, is generally a preprocessing stage that  produces an enhanced version of the document, in order to improve the recognition results of HTR engines \cite{jemni2019out}.

Early methods known by global binarization, aimed to find a single threshold value for the whole document.  A more sophisticated approaches, named local binarization, determine a different threshold value for each pixel. These threshold values are further used to classify the image pixels into foreground (black) or background (white). \cite{otsu1979threshold,sauvola2000adaptive}. 
Nowadays, thresholding based methods are still evolving, for instance a global threshold selection method was introduced in \cite{annabestani2019new} basing on fuzzy expert systems (FESs). In this method  the image contrast is enhanced in a first step. Then, a pixel-counting algorithm is used with another FES for thresholds adjustment as a range,  before choosing the right value  from  the middle of that produced list. Moreover, a support vector machine (SVM) based   approach was proposed in \cite{xiong2018degraded}. Given the local features of the gray level images, degraded regions  were classified according to SVM to be binarized according to one of four different threshold values.  
The main drawback of these thresholding methods is the sensitivity to the document condition, they usually fail to restore highly degraded images \cite{Pratikakis2018icfhr}, since it is hard to obtain a good filter in those scenarios. 

Later, energy based methods were introduced to track the text while binarizing the image. In \cite{hedjam2014constrained}, the ink was considered as a target to be maximized by an energy function, while minimizing the degraded background.  Similarly, the background is estimated and subtracted from the degraded image  using a  mathematical morphology in  \cite{xiong2018historical}. 
However, the results using those hand crafted image processing algorithms were unsatisfactory. 

Recently, deep learning architectures were used to tackle this problem by training their weights directly from raw data.  In \cite{afzal2015document},  the problem was formulated  as pixels classification depending on sequences. Hence, a  2D Long Short-Term Memory (LSTM) was used to predict each pixel value whether belonging to the text or the degradation given a sequence of its neighbours.  This process is, of course, time consuming. Thus, instead of classifying each pixel separately, images were mapped  in an end to end fashion from the degraded version into the clean one using the Convolutional Neural Networks (CNNs). These architectures, called auto-encoders, lead to recent improvements in image denoising \cite{mao2016image} and more particularly documents binarization \cite{lore2017llnet,calvo2019selectional,akbari2020binarization} or deblurring problems \cite{hradivs2015convolutional}. This kind of applications are now called image-to-image translation, since the goal is to start from a degraded image and learn a mapping function that translate it into a clean domain.  Following these approaches,   \cite{kang2021complex} proposed an auto-encoder architecture that performs a cascade of pre-trained U-Net  models \cite{ronneberger2015u} to learn the binarization with less data. Also, \cite{He2019} proposed a neural
network to learn the the enhancement/binarizatoin   in an iterative fashion.

Other deep learning approaches modeled the problem as a generation task, where the goal is to generate a clean version of the image by conditioning on the degraded one. This process was carried out using GANs architectures, composed of  a generative model that produce a clean version of the image and a discriminator to assess the binarization result. Thus,  and motivated by other approaches where GANs significantly surpasses autoencoders in  image-to-image tasks \cite{isola2017image}, some approaches applying this method were introduced.   In \cite{souibgui2020gan},
a conditional GAN approach was developed for document enhancement and achieved good results in recovering handwritten documents with several backgrounds degradation scenarios, it was also used for optical documents deblurring and dense watermarks removal. A similar cGAN based method was also proposed in \cite{zhao2019document}, where the binarization performed by the generator was done in two stages, learning the pixels in different scales then combining the results to provide the final output.  In \cite{Dang2021Binarization}, a strokes preservation method was developed using a GAN model, this was done by learning the text boundary in an auxiliary task for a better document binarization, especially with weak or ambiguous strokes. 
Another GAN's based method was proposed in \cite{bhunia2019improving} using a two  networks frameworks, for document binarization. The first one was conditioning on the clean image to generate a degraded one, while the other network reconstructed the clean version conditioning on the degraded image. Thus, an unpaired data training was performed leading to good results when using the second network to binarize images. Similarly, \cite{tamrin2021} treated the problem as two stages:  The first stage was  devoted to augment the data by creating degraded handwritten images using a GAN model, while the second stage exploited the generated images  to train an  inverse problem binarization model.

To summarize, deep learning based methods are now significantly surpassing the classic or modern image processing handcrafted algorithms for the handwritten document binarization task. However, the only limitation that can be noticed is the $"image$ $only"$  based training. Because, the usual benchmarks for testing  the  binarization performance do not include a text recognition evaluation (a GT truth text of the degraded documents). Thus, those methods could easily delete some parts of the handwritten text while binarizing the image, without noticing. {It is to note that we addressed this problem in a previous work \cite{souibgui2020conditional}, but for printed documents domain and using the Tesseract OCR engine to evaluate the produced  text. Where the character error rate of the OCR on the generated image is inputted  as an additional input channel to the discriminator.
Contrary, we are proposing in this study, is a handwritten text recognizer  that is jointly trained in the  GAN architecture to maintain the text, while cleaning the degraded image. Thus, it is more flexible to be used for different handwritten languages and writing styles.}

\begin{figure}[t]
\begin{center}
  \includegraphics[width=1.0\linewidth]{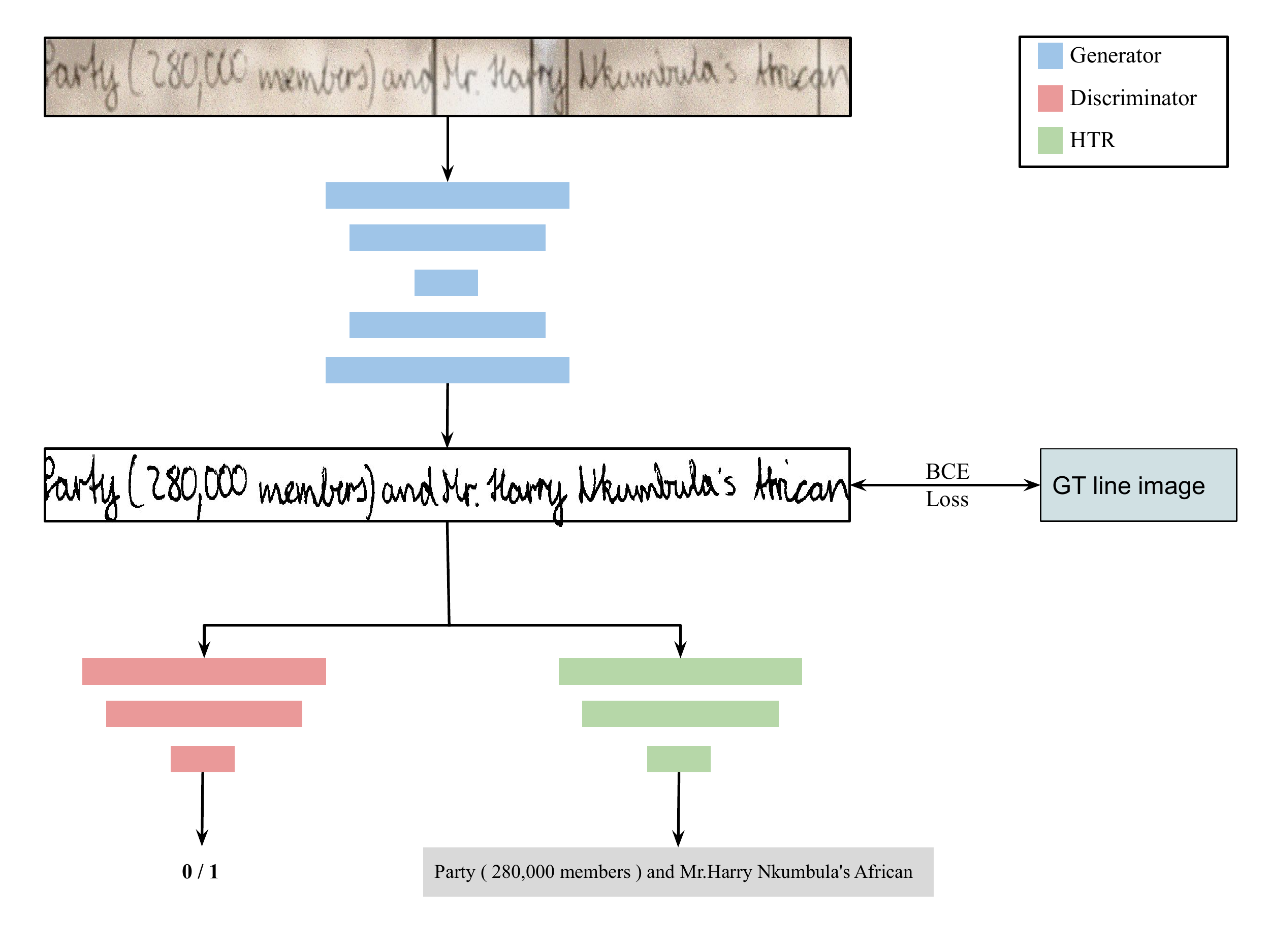}
\caption{Proposed architecture for document binarization.}
\label{fig:architecture}       
\end{center}
\end{figure}

\section{Proposed Method}\label{section:proposedapproach}

\begin{figure}[t]
\begin{center}
  \includegraphics[width=1.0\linewidth]{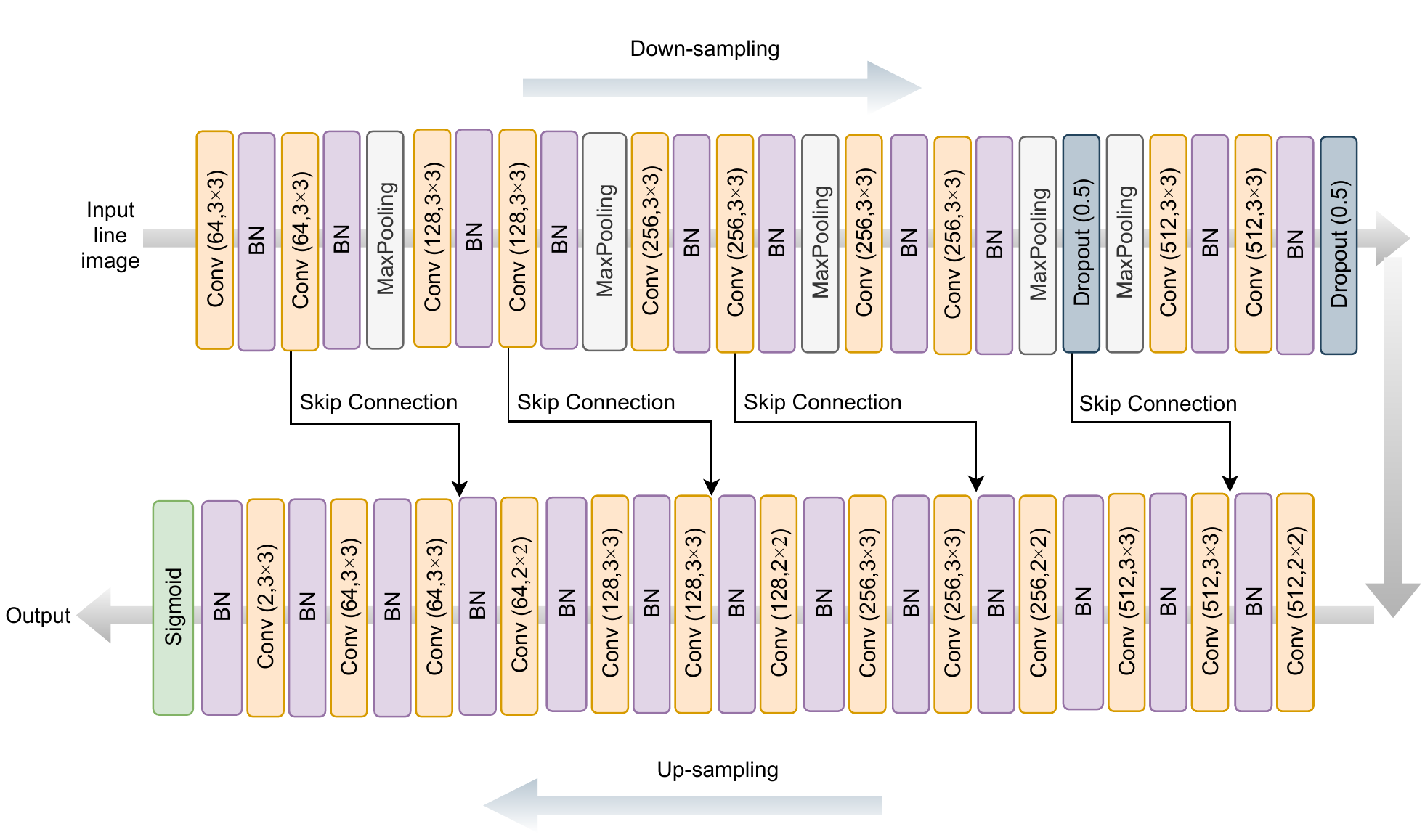}
\caption{Generator’s architecture design used in this study.}
\label{fig:generator}       
\end{center}
\end{figure}

We treat recovering a clean version from a handwritten degraded document as an image-to-image translation task using a generative model. Our GAN architecture is composed of three main parts, as shown in Figure~\ref{fig:architecture}: A generator, a discriminator and a handwritten text recognizer. Since we are using the text information, the patches that are used during training should be in a readable form  by a HTR, after binarization. Thus, the model is designed to be working at  handwritten line images level. During training,   the generator is conditioning on the degraded  line image to generate the clean version. The generated image is passed to the discriminator to assess it as real (looks clean) or fake (looks degraded), for  ensuring a realistic visually recovery. The image is also passed also to the HTR model to read it and compare the recognized text to the GT, hence, to maintain its readability while recovering it. The discriminator, as well as the recognizer, passed their feedback about the generated image through the adversarial loss.  Noting that  another  additional Binary Cross Entropy (BCE) loss is integrated in the generator, for a faster convergence. In this way,  the generator parameters are learned to produce a handwritten image that is as clean as possible, while keeping  the text quality. In what follows, we explain the three components presented in our architecture with more details.



\subsection{Generator}
Since we are doing an image-to-image translation process, the generator is  designed as an auto-encoder  model. We employ the U-net \cite{ronneberger2015u} for this task, in which  the inputted image is encoded through a sequence of convolution layers with a down-sampling to reach a specific layer. After that, the image is decoded with a sequence of up convolution  layers with an up-sampling. The model involves some skip connections after each two successive layers  to recover images with lower deterioration, since the goal is to keep the text  while removing the degradation. Thus, skip connections can help the decoder  in maintaining the text features while producing the image.     Figure~\ref{fig:generator} shows further details about the used generator. As can be seen, it is composed of  23 convolutional layers, with  Dropout regularization and batch normalization layers. The output of this model is a single channel  (in gray level) image, assumed to be the cleaned version of the inputted degraded image.


\subsection{Discriminator}

\begin{figure}[t]
\begin{center}
  \includegraphics[width=0.7\linewidth]{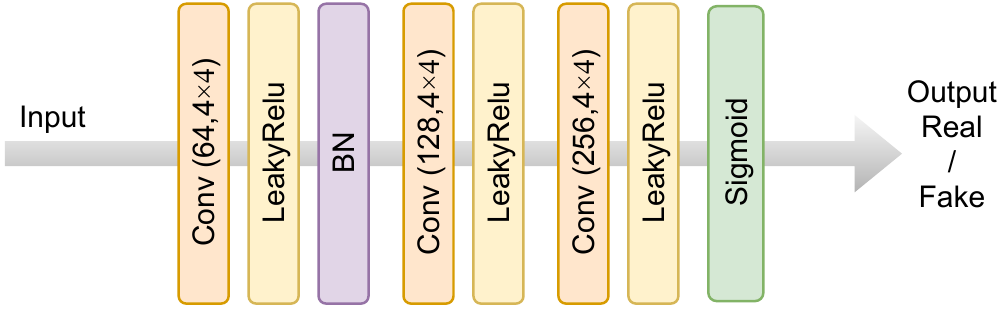}
\caption{Discriminator’s architecture used in this study.}
\label{fig:discriminator}       
\end{center}
\end{figure}

The discriminator is another Fully Convolutional Network (FCN) that produces an assessment of the generated image in term of visual similarity (pixel level) with the GT (real) images. The model was designed to take a degraded image with its clean version and output the  class "real" if the clean version is the real GT, or assign the class "fake" if the clean image was produced by the generator. Both input images, which have of course the same size, are  concatenated in an $H\times W \times2$ shape. Then, the obtained volume is propagated in the discriminator  model detailed in Figure~\ref{fig:discriminator}, to end up in the last layer as a form of   $H/16 \times W/16 \times 1$ matrix. During training, this matrix contains values that are equal to 1 in  case of inputting the GT as a clean image, and equal to 0 in  case of inputting the generator based enhanced image.


\subsection{Handwritten Text Line Recognizer}

\begin{figure}[t]
\begin{center}
  \includegraphics[width=1.0\linewidth]{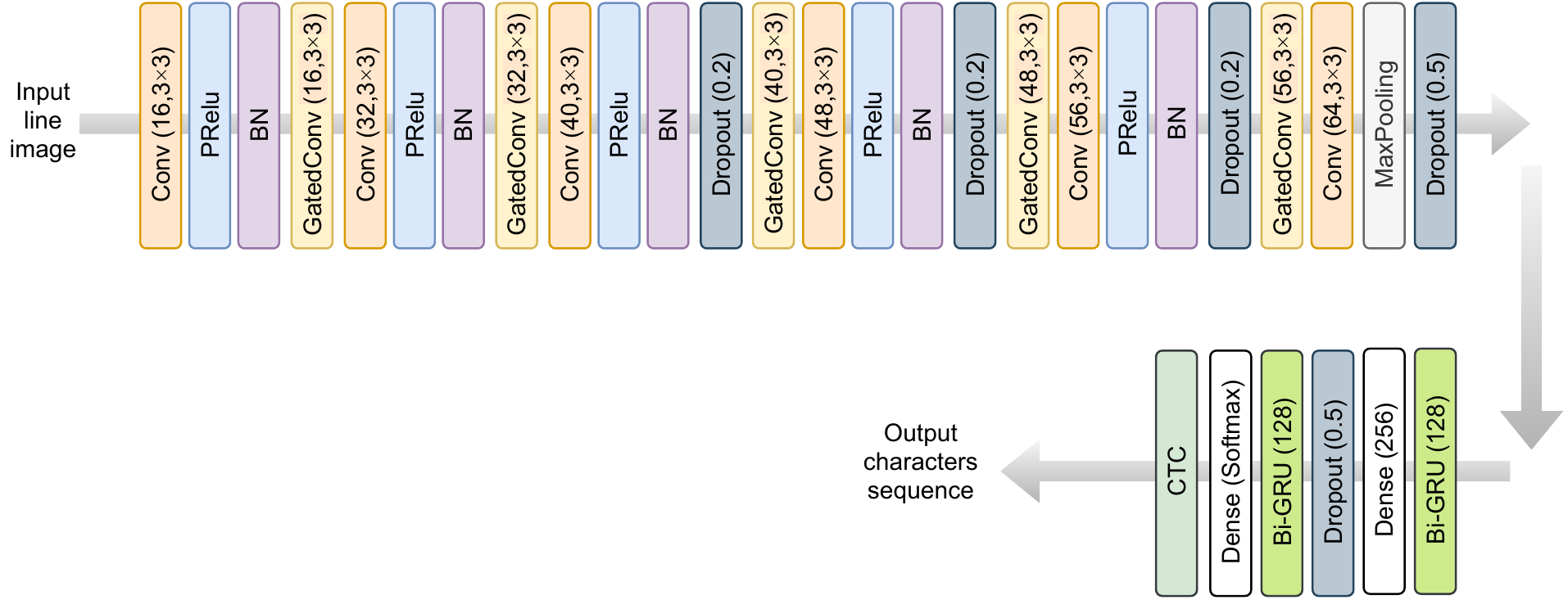}
\caption{Workflow of the CNN-Bi-GRU recognizer’s architecture.}
\label{fig:recognizer_architecture}       
\end{center}
\end{figure}

The used handwritten recognizer is a Convolutional Recurrent Neural Network (CRNN) model,  following the  architecture presented recently in \cite{flor2020CRNN} and considered among the best HTR architectures.  Noting that, any other HTR  can be also used for this task, for instance:  \cite{bluche2017ICDAR, jemni2018offline, kang2021candidate}.
The model architecture is detailed in Figure~\ref{fig:recognizer_architecture}.   After enhancing the image by the generator it is inputted  to an encoding stage  that uses  convolutional and gated convolutional layers, with integrated regularization techniques. The encoded image is passed later to the decoding stage, consists in two bidirectional  Gated Recurrent Unit (GRU) layers.  Finally,  the CTC  is used to decode the feature frames into text characters. The CTC  layer is having the size of the character set plus one additional symbol corresponding to the blank symbol.   During training, the recognizer could be fitted with two types of clean images, forming the following  two scenarios:

\begin{itemize}
    \item \textbf{S1:} The Recognizer is trained at each iteration with the GT images that are related to the degraded batch images inputted to the generator. The GT images are used with the associated GT text transcription for training in this process. 
    
    \item \textbf{S2:} The Recognizer is trained using the images enhanced by the generator at each iteration with the associated GT text. The intuition behind, is that we may obtain a better recognition convergence that is going progressively from the degraded domain to the clean domain.   
\end{itemize}

\subsection{Training process}

The different components presented above were trained  jointly. The generator $G$, which is a function having  the parameters $\theta_G$, is conditioned on the degraded image $I_d$ to provide its  cleaned image that should be as close as possible to the GT image $I_{gt}$.  This image is passed to be validated by the discriminator $D$ and handwritten recognizer $R$, with parameters $\theta_D$ and $\theta_R$, respectively.  $D$ is giving an assessment of  the cleaned image about its cleanliness to be Real or Fake, $P(\text{Real}) = D_{\theta_D} (G_{\theta_G}(I_d))$. This  adversarial training process of $G$ and $D$ can be formalized by:

\begin{dmath}
\mathcal{L}_{adv}(\theta _G,\theta _D) = {\mathbb{E}_{{I_d},{I_{gt}}}}\log [{D_{{\theta_D}}}({I_d},{I_{gt}})] + {\mathbb{E}_{{I_d}}} \log [1 - {D_{{\theta _D}}}({I_d},G_{\theta_G}(I_d))] 
\end{dmath}

 $R$ is recognizing the generated image to maintain its readability with the CTC decoder, $CTC(t,{R_{{\theta_R}}}({G_{{\theta_G}}}({I_d})))$, where $t$ is the GT text. Note that it is trained with a clean version of the image (whether using S1 or S2, presented above),  at each same iteration: $CTC(t,{R_{\theta_R}}(I_{gt}))$. Also, for a faster convergence, a simple BCE loss is used in the generator  between the cleaned images and the GT ones,  $BCE(\theta_G)$. Thus, the generator is being affected by three factors to produce its generation. The whole architecture is formalized as: 

\begin{dmath}\label{eq:full_loss}
    \mathcal{L}(\theta _G,\theta _D,\theta _R) =  min_{\theta_G} max_{\theta_D} L_{adv}(\theta_G,\theta_D) +  \lambda ({E_{t,{I_d}}} CTC(t,{R_{{\theta_R}}}({G_{{\theta_G}}}({I_d}))) + \beta BCE(\theta_G) 
\end{dmath}

Where $\lambda$ and $\beta$ are the weights balancing the components intervention to produce the final generated image. During our experiments, we set $\lambda$ to 1 and $\beta$ to 10.  For training, we used Adam’s optimizer  for the generator and discriminator components, while using the   RMSProp  
for the handwritten text recognizer.

\section{Experiments and Results}\label{section:experiments}
We provide in this Section the experiments that were done to validate the effectiveness of our proposed method. First, we start by presenting the metrics and datasets used in our evaluation. 




\subsection{Metrics}
Following the usual approaches for handwritten document image binarization \cite{Pratikakis2012ICFHR}, we use the same metrics to validate the cleaned images. Those metrics which compare the images visual similarity with the GT clean ones, are: Peak signal-to-noise ratio (PSNR), F-Measure (FM), pseudo-F-measure (Fps) and Distance reciprocal distortion metric (DRD).
In addition, since we are using the text information to validate our model, we utilise as well the HTR metrics for comparing the recognized text to the GT one. These metrics are based on the Levenshtein distance \cite{levenshtein1966binary}, and they consists in the  Character Error Rate (CER) and the Word Error Rate (WER) measures.

\subsection{Handwritten text databases}
Usual handwritten document binarization databases do not contain text information \cite{Pratikakis2018icfhr,burie2016icfhr2016,ayatollahi2013persian}. Thus, we opt to create synthetically degraded images from the databases used in HTR tasks in order to exploit the GT text provided within these datasets.  
We address in this study  two different  alphabets: Arabic and Latin. From each, we took the most used database for handwritten text line image recognition: KHATT \cite{mahmoud2014khatt} and IAM \cite{marti2002iam}, to add degradation. We call the created datasets, degraded-KHATT and degraded-IAM.

\subsubsection{Degraded-KHATT}
\begin{figure}[h]
\begin{center}

\begin{tabular}{c}

\noalign{\smallskip}

\multicolumn{1}{l}{\includegraphics[width=0.8\columnwidth, height=0.6cm]{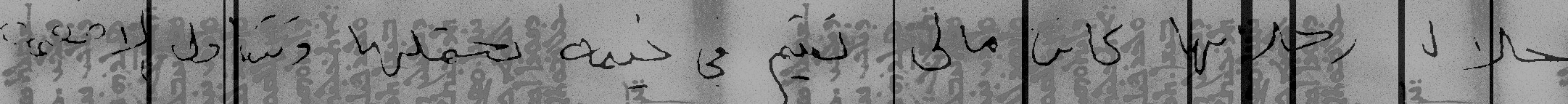}}\\

\noalign{\smallskip}

\multicolumn{1}{l}{\includegraphics[width=0.8\columnwidth, height=0.6cm]{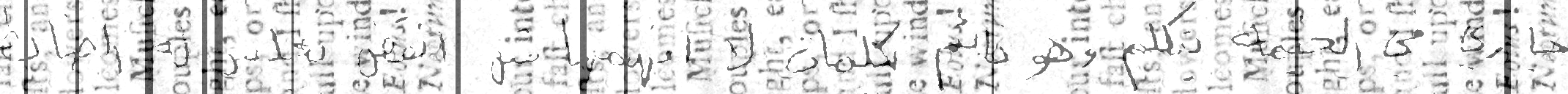}}\\

\noalign{\smallskip}

\multicolumn{1}{l}{\includegraphics[width=0.8\columnwidth, height=0.6cm]{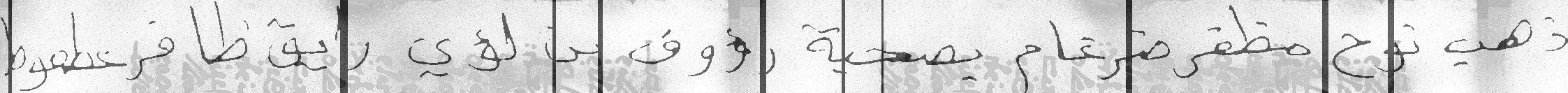}}\\

\noalign{\smallskip}
\multicolumn{1}{l}{\includegraphics[width=0.8\columnwidth, height=0.6cm]{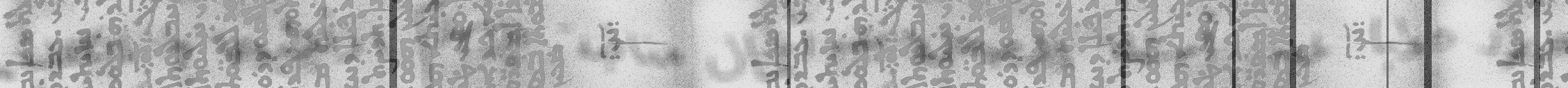}}\\

\noalign{\smallskip}
\multicolumn{1}{l}{\includegraphics[width=0.8\columnwidth, height=0.6cm]{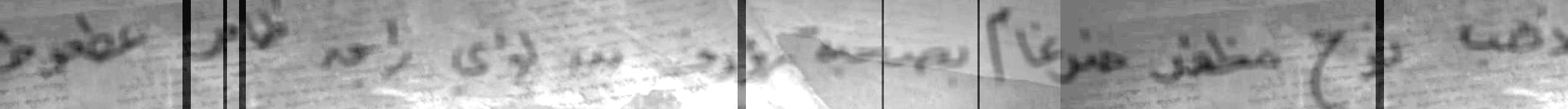}}\\

\noalign{\smallskip}

\multicolumn{1}{l}{\includegraphics[width=0.8\columnwidth, height=0.6cm]{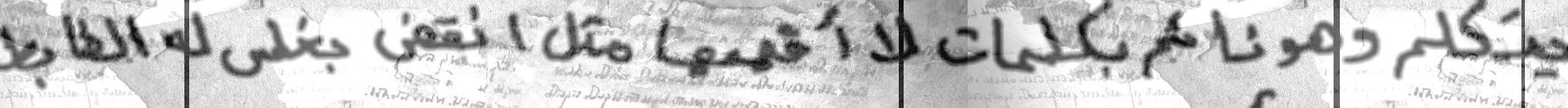}}\\

\noalign{\smallskip}

\multicolumn{1}{l}{\includegraphics[width=0.8\columnwidth, height=0.6cm]{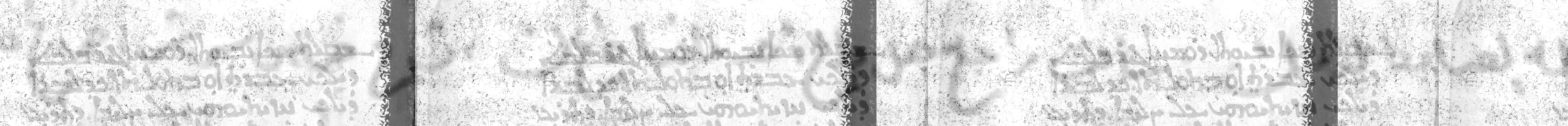}}\\

\noalign{\smallskip}
\end{tabular}

\caption{Examples of distorted line images  of the degraded-KHATT database used in this study, images are presented in gray level.}
\label{fig:distorted_khatt} 
\end{center}
\end{figure}

The KHATT dataset was developed for Arabic manuscript recognition, contains texts lines images with their associated GT texts.  In our experiments, we used 6161 
lines for training and a set of 1861 lines for testing, while a set of 940 lines 
was used for validation as it was done  in  \cite{bluche2017ICDAR}. Then, we  added random  distortions as shown in 
Figure~\ref{fig:distorted_khatt} to obtain the degraded-KHATT dataset. {To accomplish this, we insert different background images containing some flaws or artifacts. These background images are extracted mainly from public historical documents such as Nabuco, Bickley diary and Persian datasets. We have also applied different distortion operations, especially, dilation, erosion and blurs using random kernel sizes ($2\times2$  and $3\times3$ for dilation,  $2\times2$ , $3\times3$ and $4\times4$ for erosion and from $1\times1$ to  $15\times15$ for blurring). We inserted also random vertical lines having random widths in order to simulate the noise that can occur in old historical documents.}

\subsubsection{Degraded-IAM}

\begin{figure}[h]

\begin{center}

\begin{tabular}{c}

\noalign{\smallskip}
\multicolumn{1}{l}{\includegraphics[width=0.8\columnwidth, height=0.6cm]{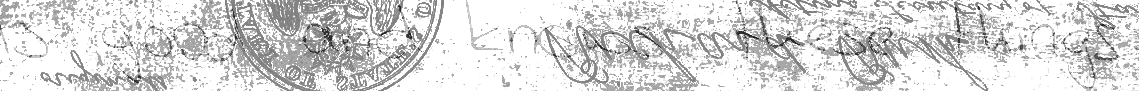}}\\
\noalign{\smallskip}
\multicolumn{1}{l}{\includegraphics[width=0.8\columnwidth, height=0.6cm]{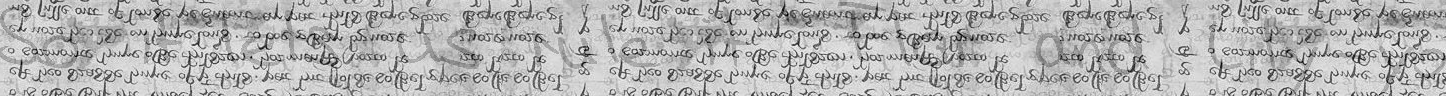}}\\
\noalign{\smallskip}
\multicolumn{1}{l}{\includegraphics[width=0.8\columnwidth, height=0.6cm]{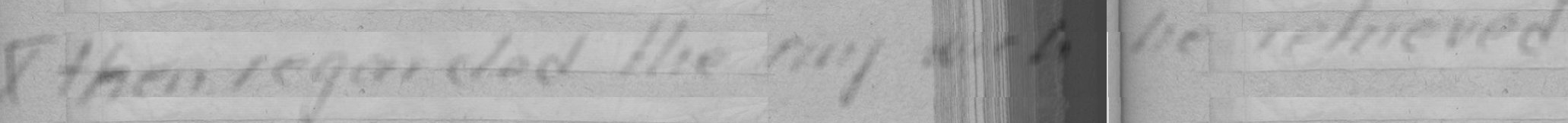}}\\
\noalign{\smallskip}
\multicolumn{1}{l}{\includegraphics[width=0.8\columnwidth, height=0.6cm]{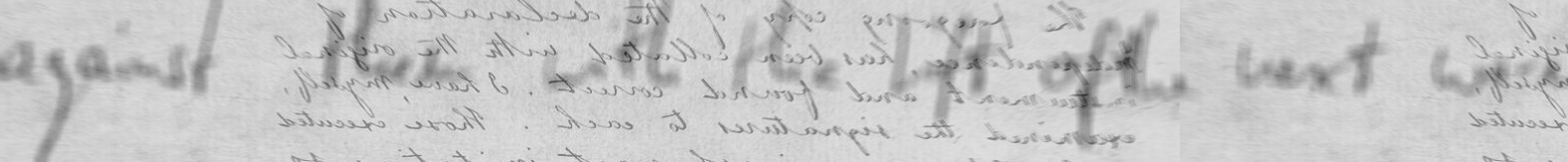}}\\
\noalign{\smallskip}
\multicolumn{1}{l}{\includegraphics[width=0.8\columnwidth, height=0.6cm]{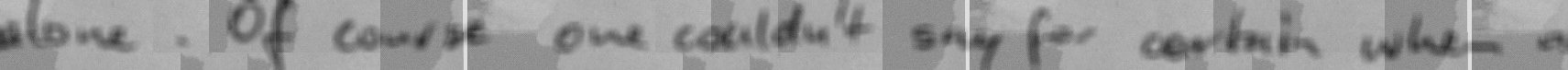}}\\

\end{tabular}

\caption{Examples of distorted line images 
of the degraded-IAM database used in this study, images are presented in gray level.}
\label{fig:distorted_iam} 
\end{center}
\end{figure}

The IAM dataset was proposed for handwritten Latin script text recognition. It contains 8962 line images taken from the  Lancaster-Oslo/Bergen (LOB) corpus. To insert degradation, we used same as in KHATT, 6161, 940, 1861 line images for training, validation and testing, respectively. We add dense backgrounds to simulate real historical deteriorated images  same as  it was done for the degraded-KHATT presented above. Examples of the obtained degraded-IAM are illustrated in Figure~\ref{fig:distorted_iam}. 


\subsection{Results}
\subsubsection{Arabic handwritten texts images recovery}

For Arabic, we fed the proposed model with the training set of the created degraded-KHATT database. As stated above, the generator is trained to map the degraded image into a clean version, which will be evaluated by the discriminator and the recognizer. It is to note that in our experiments (for Arabic and Latin manuscripts), we used a high degradation for a  meaningful evaluation in the hard scenarios. Also, we separate the background types between the training and testing sets (i.e there is no intersection in the background noise between the two sets).

\begin{table*}[!b]

\caption{{Image binarization results for the \textit{test set} (degraded-KHATT database). (A	$\rightarrow$ B): The CRNN is trained on images from the domain A and tested on images from the domain B. Deg.: Degraded images. Reco.: Recognition performance. }}\label{tab_enhance_test_khatt}
\centering
\scriptsize
\begin{tabular}{|c||c|c|c|c||c|c|c|c|}
\noalign{\smallskip}
\hline
             & \multicolumn{4}{c||}{\vtop{\hbox{\strut Binarization Performance}\hbox{\strut (Visual Quality)}}}                                                                                          & \multicolumn{2}{c|}{Reco. CRNN1 \%} & \multicolumn{2}{c|}{Reco. CRNN2 \%}  \\ \cline{2-9} 
{Method} & PSNR                          & FM                     & Fps                           & DRD                          & CER                         & WER   & CER                         & WER \\ \hline\hline

\vtop{\hbox{\strut CRNN \cite{flor2020CRNN}}\hbox{\strut (GT 	$\rightarrow$ GT)}}             & ND                            & ND                            & ND                            & ND                           & 12.04                         & 32.39      & -                         & -              \\ \hline

\vtop{\hbox{\strut CRNN \cite{flor2020CRNN}}\hbox{\strut (Deg. 	$\rightarrow$ Deg.)}}& 4.80 &	25.45 &	25.70 &	107.22 & 30.34  & 54.44  & -                         & - \\ \hline

\vtop{\hbox{\strut CRNN \cite{flor2020CRNN}}\hbox{\strut (GT 	$\rightarrow$ Deg.)}}& 4.80 &	25.45 &	25.70 &	107.22  & 91.18 & 100      & -                         & -               \\ \hline\hline

Baseline cGAN          & \textbf{\textbf{15.52}} &	75.01 &	75.11 &	\textbf{\textbf{6.05}}  & 29.24 & 53.68                      & -                         & -  \\ \hline
cGAN \cite{souibgui2020conditional} 

    & 15.10 &	75.56	& 75.75 & 11.78 & 28.84 & 54.37                  & -                         & -         \\ \hline
 \textbf{Ours  (S1)}   & 15.45	& \textbf{\textbf{77.45}} &	\textbf{\textbf{77.60}} &	7.97   &   27.03    &  52.84      &     \textbf{\textbf{24.33}}                       &   \textbf{\textbf{47.67}} \\ \hline
 
  \textbf{Ours  (S2)}  & 15.44	& 74.52	& 74.62	& 6.18    &   27.90  &    53.49  &             25.31              & 48.48  \\ \hline
 
\end{tabular}
\end{table*} 
 
{Table~\ref{tab_enhance_test_khatt} illustrates the obtained results of the performed image binarization methods on the test set of the degraded-KHATT database. Reminding that we proposed two binarization scenarios sharing the same GAN architecture and integrating a CRNN recognizer, S1 and S2, stated before. The key difference between the two scenarios is the data fed to the recognizer during the GAN training stage. In scenario S1, the recognizer (we call it CRNN1) is fed with ground truth clean images, while it is fed with generated images (cleaned by the generator) in the second scenario (called CRNN2).
}
As it can be seen, contrary to the previous related approaches, we evaluate the image in its {visual quality (Binarization performance)} and readability (recognition performance) at the same time. For  readability, we tested each of our both  scenarios S1 and S2 {(Reco. CRNN1 and Reco. CRNN2) using the two recognizers (CRNN1 and CRNN2)}. 

To compare our approach, we used a simple GAN architecture as a baseline. The architecture contains the same generator and discriminator of our architecture, but, without using a recognizer. We compare also with the method presented in \cite{souibgui2020conditional} for printed text recovery, where an OCR is used during training as a part of the discriminator. However, since we are doing a handwritten text recognition (not optical text). We modify it by training a HTR having the same architecture as \cite{flor2020CRNN} to use it as a part of the discriminator, more details are given in \cite{souibgui2020conditional}.


 \begin{figure}[t]
\centering
\begin{tabular}{l@{\hskip 0.1in}l}
\hline
 \noalign{\smallskip}
  (a): &\includegraphics[width=85mm, height=4.5mm]{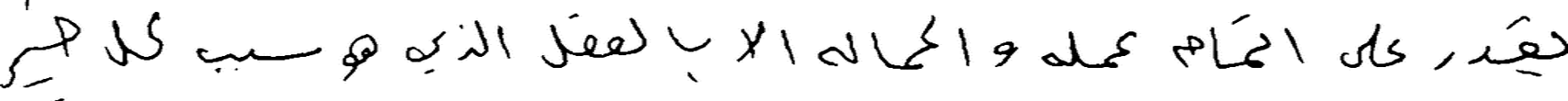}
 \\[2mm]  
   (b): & \includegraphics[width=85mm, height=4.5mm]{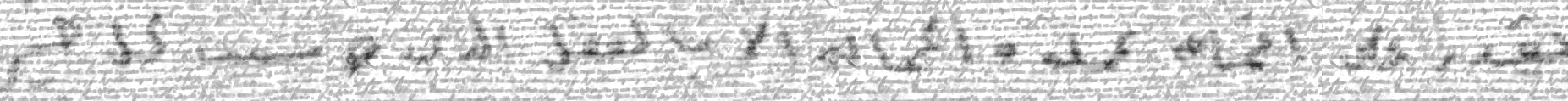}
 \\[2mm]  
   (c): &\includegraphics[width=85mm, height=4.5mm]{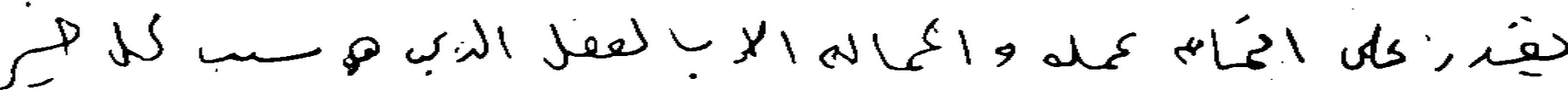}
 \\[2mm]  
(d):  &\includegraphics[width=85mm, height=4.5mm]{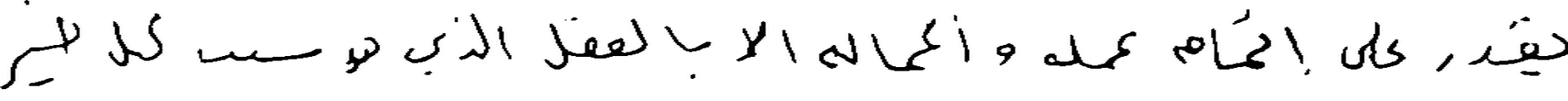}
 \\[2mm]  
    (e): &\includegraphics[width=85mm, height=4.5mm]{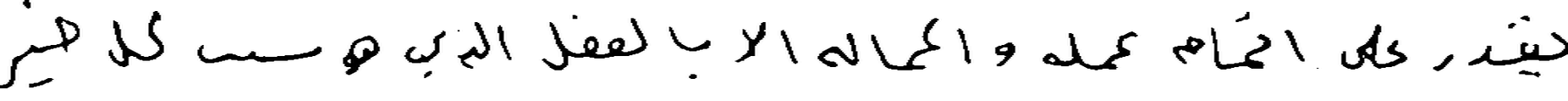}
 \\[2mm]  
     (f): &\includegraphics[width=85mm, height=4.5mm]{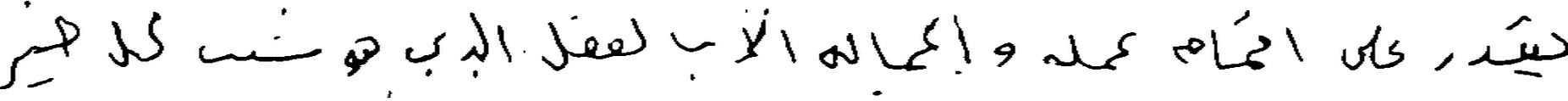}
 \\[2mm]  
 \hline
 \noalign{\smallskip}

   (a): &\includegraphics[width=85mm, height=4.5mm]{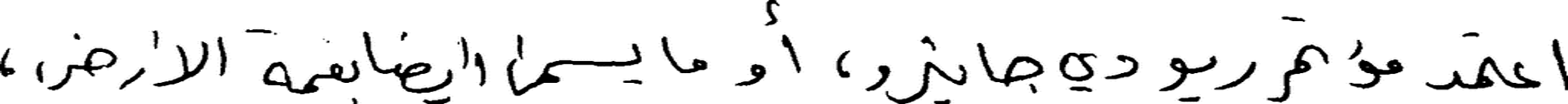}
 \\[2mm]  
   (b): & \includegraphics[width=85mm, height=4.5mm]{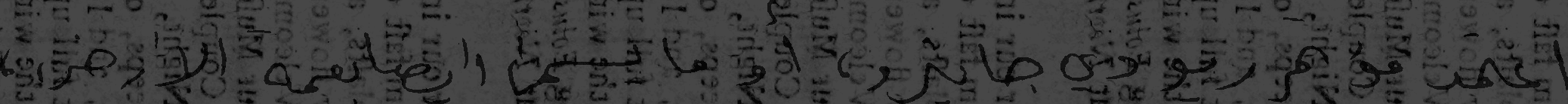}
 \\[2mm]  
   (c): &\includegraphics[width=85mm, height=4.5mm]{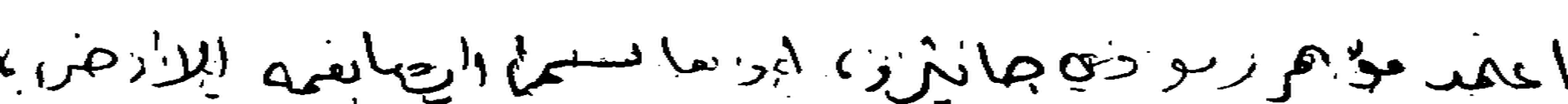}
  \\[2mm]  
(d): &\includegraphics[width=85mm, height=4.5mm]{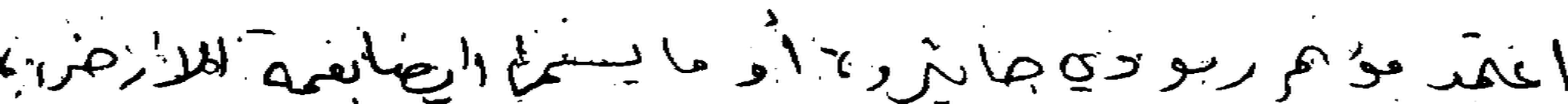}
 \\[2mm]  
     (e): &\includegraphics[width=85mm, height=4.5mm]{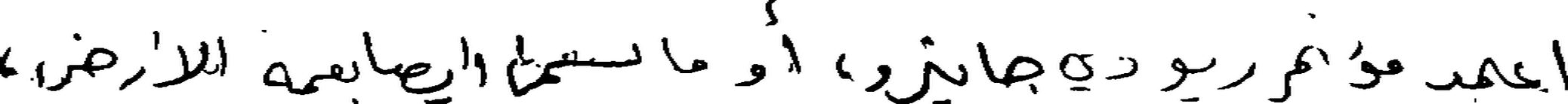}
 \\[2mm]  
     (f): &\includegraphics[width=85mm, height=4.5mm]{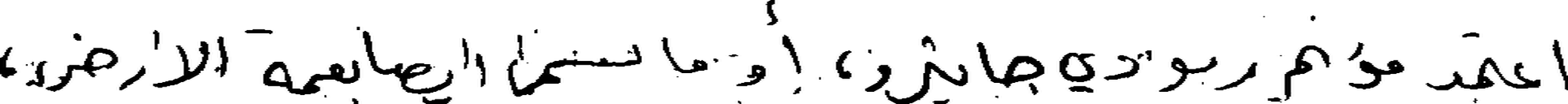}
 \\[2mm]  
 \hline
 \noalign{\smallskip} 
 \end{tabular}
\caption{Results of our proposed method for recovering degraded lines images. (a): GT, (b): Distorted, (c): Baseline cGAN, (d): cGAN \cite{souibgui2020conditional}, (e): Ours S1, (f): Ours S2.}
\label{fig:khatt_recovering}
\end{figure}

Out of the results, we can see that using the GT  images, a trained HTR engine based on CRNN \cite{flor2020CRNN} is reaching a CER of 12.04 \% and 32.39 \% as a  WER, this is considered as our upper bound for recognition. Using the same  model to recognize the degraded images, we obtain a poor performance of 91.18 \% in CER, obviously because the model is trained on the clean data. If we train the model on the degraded train set, it results in  30.34 \% of CER and  54.44 \% as WER. This experiment is done to verify later if we can surpass this performance (as a baseline) by cleaning the images then reading them, instead of training a model on the degraded domain.

\begin{figure}[htb!]
\centering
\begin{tabular}{l@{\hskip 0.1in}l}
\hline
 
 \noalign{\smallskip}
 
   (a): &\includegraphics[width=85mm, height=4.5mm]{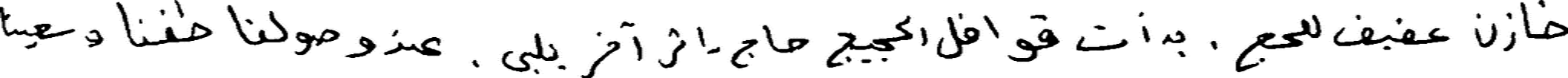}
 \\[2mm]  
   (b): & \includegraphics[width=85mm, height=4.5mm]{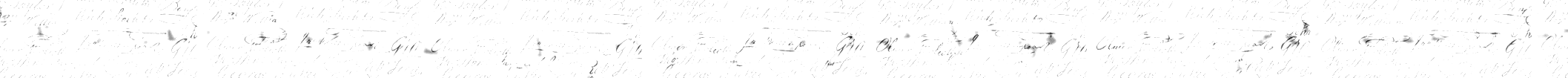}
 \\[2mm]  
  (c): &\includegraphics[width=85mm, height=4.5mm]{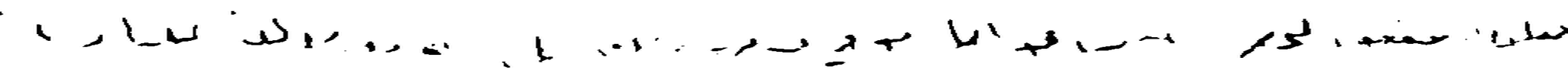}
   \\[2mm]  
(d): &\includegraphics[width=85mm, height=4.5mm]{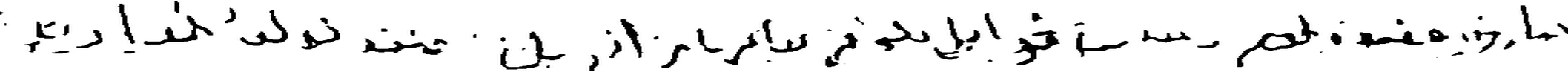}
 \\[2mm]  
     (e): &\includegraphics[width=85mm, height=4.5mm]{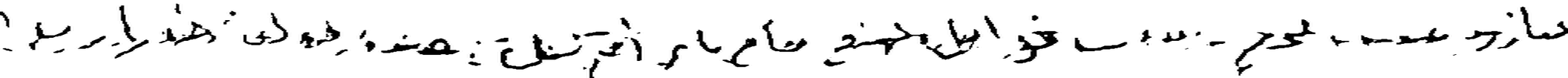}
 \\[2mm]  
     (f): &\includegraphics[width=85mm, height=4.5mm]{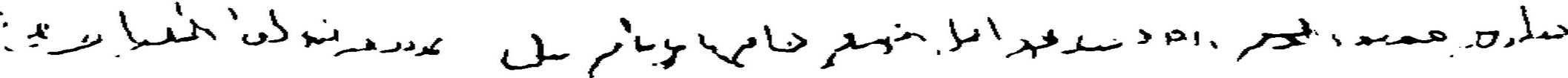}
 \\[2mm]  
 \hline
 \noalign{\smallskip}
 
   (a): &\includegraphics[width=85mm, height=4.5mm]{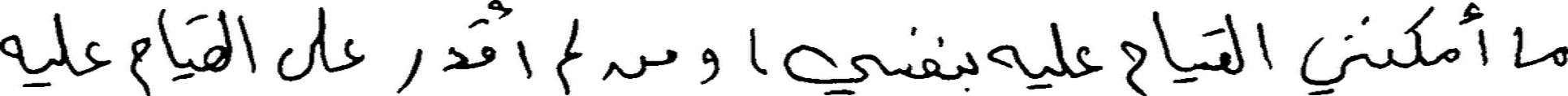}
 \\[2mm]  
   (b): & \includegraphics[width=85mm, height=4.5mm]{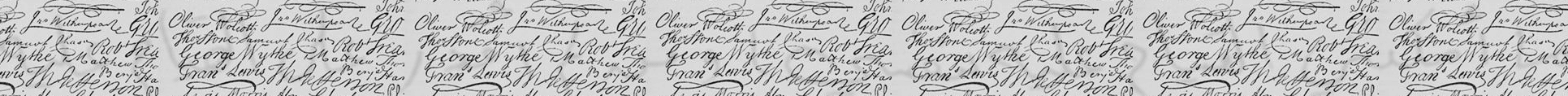}
 \\[2mm]  
  (c): &\includegraphics[width=85mm, height=4.5mm]{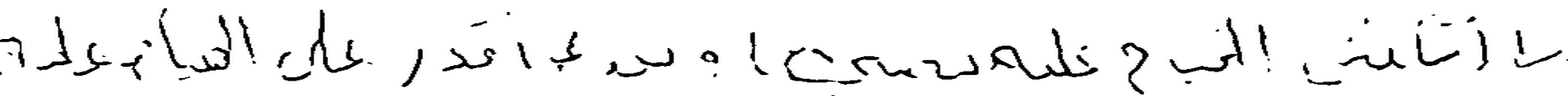}
   \\[2mm]  
(d): &\includegraphics[width=85mm, height=4.5mm]{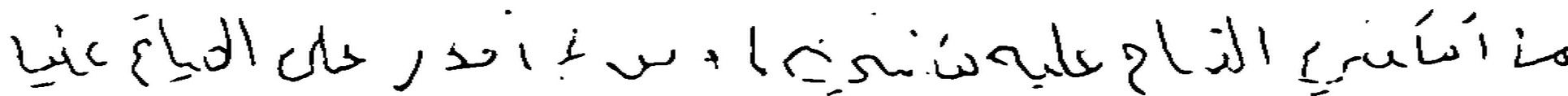}
 \\[2mm]  
     (e): &\includegraphics[width=85mm, height=4.5mm]{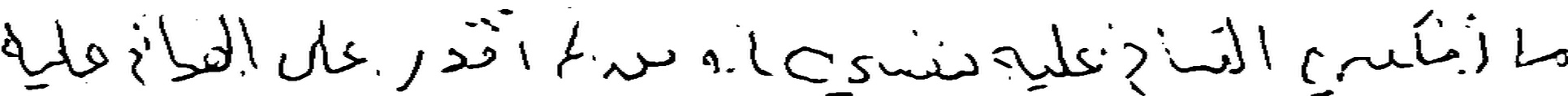}
 \\[2mm]  
     (f): &\includegraphics[width=85mm, height=4.5mm]{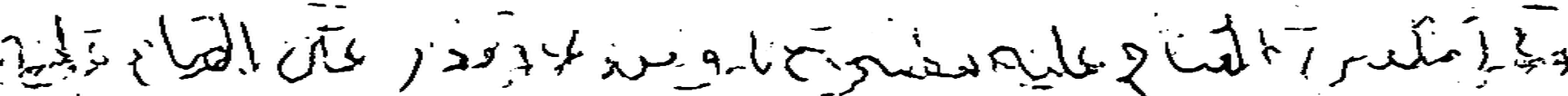}
 \\[2mm]  
 \hline
 \noalign{\smallskip} 
 
\end{tabular}
\caption{Results of our proposed method for recovering extremely degraded lines images.  (a): GT, (b): Distorted, (c): Baseline cGAN, (d): cGAN \cite{souibgui2020conditional}, (e): Ours S1, (f): Ours S2.}
\label{fig:khatt_recovering_extremely}
\end{figure}

The different binarization approaches, as it can be noticed, are enhancing the visual quality and the readability of the degraded lines. However, we can see that the baseline cGAN which is not taking the text into consideration while cleaning the image, is producing a result having a better visual quality in term of PSNR and DRD, but a worse readability compared to the methods integrating the text information. { For the recognizer based methods, it is clear that our recovery method (S1) is leading to the best performance in terms of having a good visual enhancement while conserving the text readability}. Since by recognizing the produced images, we get a CER of 27.03 \% and a WER of 52.84 \% when using the recognizer of S1 and a CER of 24.33 \% and a WER of 47.67 \% when using the S2 recognizer. {This proves that using the text information during binarizing the images is useful. Also, we notice that the progressive learning of a HTR (training in an order from the degraded images to their clean versions) in a multitask framework,  is better for the recognition task. However,  using the HTR pretrained with the clean GT images (as a separate task) during enhancing the document,  is better for the binarization performance (visually).}  
To illustrate our method's effectiveness, we show in Figure~\ref{fig:khatt_recovering} and Figure~\ref{fig:khatt_recovering_extremely} some qualitative results of recovering the distorted lines, ranging from easy distortions to the hard ones. We can see that our method is the most successful in producing clean images especially in cases of highly degraded ones. In fact, it can even recover the vanished handwritten text strokes.

\begin{table}[t]
\caption{Impact of the recognizer weight on the final generated image.}\label{tab_ablation_study}
\centering
\scriptsize
\small
\begin{tabular}{|c|c|c|c|}
\noalign{\smallskip}
\hline  
&   \vtop{\hbox{\strut Binarization Performance}\hbox{\strut (Visual Quality)}}   & \multicolumn{2}{c|}{Reco. performance}  \\
\cline{2-4} 
$\lambda$ & PSNR & CER\%            & WER\%            \\ \hline
0.5                                & \textbf{17.94}               & 11.98            & 31.07             \\ \hline

\textbf{1}                                  & {17.88}                & \textbf{11.74}            & \textbf{31.05}            \\ \hline
5                                  & 17.71                  & 12.66            & 32.44            \\ \hline
10                                 & 17.07                  & 15.22            & 36.74            \\ \hline
20                                 & 16.32                  & 19.72            & 40.84            \\ \hline
\end{tabular}
\end{table}


Furthermore, as we stated above and since different weights can be used  in the recognizer loss level to control the enhancement, we perform an ablation study  to evaluate the right trade of between the visual quality and the readability during enhancing. In other words, the effectiveness of the weight $\lambda$ presented in Equation \ref{eq:full_loss}. This is done  by varying the weight $\lambda$, then  training the model with that setting and finally measuring the visual quality and readability {(using the recognizer of S1)} at each time to have the right option. The obtained results are shown in Table~\ref{tab_ablation_study}, where the experiments were carried out using the first scenario (S1) on a set of the Degraded-KHATT database, and ended up by selecting the setting of $\lambda$ to be 1.

\subsubsection{Latin handwritten texts images recovery}

For Latin manuscript, we performed the same experiments using the degraded-IAM dataset. The obtained results are presented in Table~\ref{tab:resuts_iam}. As it can be seen, training a CRNN \cite{flor2020CRNN} on the  degraded images leads to 40.34 \% as a CER and a WER of 74.05 \%, with an obvious poor visual binarization quality since there was not a performed binarization with this way (using directly the degraded version). Contrary, by cleaning the image and passing it to the recognizer, better results were obtained. Here, same as the previous experiment, we are comparing our method to the basic GAN (without a recognizer), to validate the use of text information in our current method and our proposed one in \cite{souibgui2020conditional}. It can be noticed that our method surpasses both GAN methods in the visual quality, and achieves the best text recognition rate compared to the other options.
By using the recognizer trained in S1, we boost the CER  by 1.50 \% compared to \cite{souibgui2020conditional}, 5.46 \% compared to the basic GAN and  14.29 \% compared to reading handwritten images in the distorted  domain. Moreover, using the recognizer trained in S2 we can even improve the CER result by 4.07 \%.

\begin{table}[h]
\caption{{Image binarization results for the \textit{test set} (degraded-IAM database).
(A	$\rightarrow$ B): The CRNN is trained on images from the domain A and tested on images from the domain B. Deg.: Degraded images. Reco.: Recognition performance.}}
\centering
\scriptsize
\begin{tabular}{|c||c|c|c|c||c|c|c|c|}
\noalign{\smallskip}
\hline
             & \multicolumn{4}{c||}{ \vtop{\hbox{\strut Binarization Performance}\hbox{\strut (Visual Quality)}} }                                                                                          & \multicolumn{2}{c|}{Reco. CRNN1 \%} & \multicolumn{2}{c|}{Reco. CRNN2 \%}  \\ \cline{2-9} 
{Method} & PSNR                          & FM                     & Fps                           & DRD                          & CER                         & WER   & CER                         & WER \\ \hline\hline
 \vtop{\hbox{\strut CRNN \cite{flor2020CRNN}}\hbox{\strut (GT 	$\rightarrow$ GT)}}                 & ND                            & ND                            & ND                            & ND                           & 11.92                         & 36.07  & - & -   \\ \hline

\vtop{\hbox{\strut CRNN \cite{flor2020CRNN}}\hbox{\strut (Deg. 	$\rightarrow$ Deg.)}}  & 6.01                          & 26.13                         & 26.12                         & 70.81                        & 40.34                         & 74.05                       & - & -    \\ \hline 

\vtop{\hbox{\strut CRNN \cite{flor2020CRNN}}\hbox{\strut (GT 	$\rightarrow$ Deg.)}}
       & 6.01                          & 26.13                         & 26.12                         & 70.81                        & 90.46                         & 99.50                        & - & -   \\ \hline\hline

Baseline cGAN          & 14.99                         & 75.44                            & 75.01                         & 5.91                         & 31.51 & 60.95  & - & -                                               \\ \hline

cGAN \cite{souibgui2020conditional} & 15.86 & 80.89 & 80.83  & 5.00  & 27.55 &  58.08  & - & -   \\ \hline

  \textbf{Ours (S1)}    &\textbf{\textbf{15.97}} & \textbf{\textbf{81.69}} & \textbf{\textbf{81.55}} & \textbf{\textbf{4.83}}&   26.05&  56.07   &  \textbf{\textbf{21.98}} & \textbf{\textbf{49.74}}   \\ \hline
 
   \textbf{Ours (S2)}    &  15.87 &  81.12 &	81.16 &	5.09    &   27.48 &   58.35    & 23.07 & 51.15   \\ \hline

\end{tabular}
\label{tab:resuts_iam}
\end{table}

\begin{figure}[H]
  \begin{center}
 \begin{tabular}{l@{\hskip 0.1in}l}

   GT image: &\includegraphics[width=80mm, height=5mm]{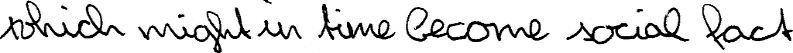}
  \\[2mm]  
   GT text: & \small which might in time become social fact
  \\[2mm]  
   R(GT): & \small which might in time become social fact
  \\[2mm]  
 
   Distorted: & \includegraphics[width=80mm, height=5mm]{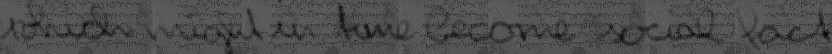}
  \\[2mm]  
   R(D): & \small  \textcolor{red}{pse nvgd} in \textcolor{red}{ha}me \textcolor{red}{Ce}come s\textcolor{red}{e}cial \textcolor{red}{ho}t
  \\[2mm]  
    R(GT): & \small  \textcolor{red}{f}
  \\[2mm]  
   Baseline: &\includegraphics[width=80mm, height=5mm]{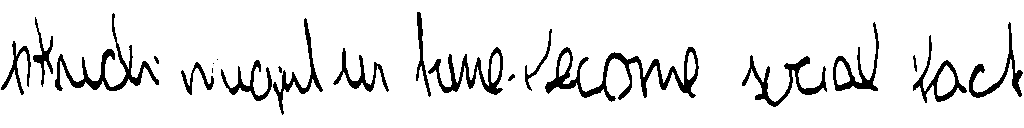}
  \\[2mm]  
   R(GT): & \small \textcolor{red}{ntauder} \textcolor{red}{m}\textcolor{red}{uornd} in \textcolor{red}{h}me \textcolor{red}{v}ecome \textcolor{red}{wu}al \textcolor{red}{t}ac\textcolor{red}{k}
  \\[2mm]

    cGAN \cite{souibgui2020conditional}: &\includegraphics[width=80mm, height=5mm]{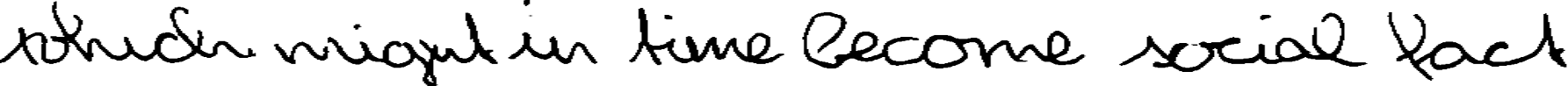}
  \\[2mm]  
   R(GT): & \small  \textcolor{red}{Anden}mig\textcolor{red}{n}t in time become social \textcolor{red}{K}ac\textcolor{red}{k}
  \\[2mm]  
  
   \textbf{Ours (S1)}: &\includegraphics[width=80mm, height=5mm]{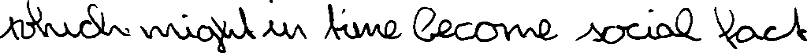}
  \\[2mm]  
   R(GT): & \small wh\textcolor{red}{u}ch\textcolor{red}{-}might in time become social \textcolor{red}{L}ac\textcolor{red}{k}
  \\[2mm]  

 R(Generated): & \small \textcolor{red}{xhud-}mig\textcolor{red}{u}t in time \textcolor{red}{G}ecome social fact
  \\[2mm]  

  Ours (S2): &\includegraphics[width=80mm, height=5mm]{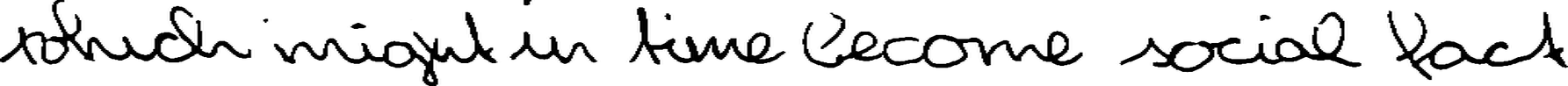}
  \\[2mm]  
  R(GT): & \small \textcolor{red}{Adu}ch\textcolor{red}{i}might in time \textcolor{red}{V}ecome social \textcolor{red}{L}act  
  \\[2mm]  
 
  R(Generated) : & \small \textcolor{red}{rdu}ch mg\textcolor{red}{u}t in time become social fact
   
   \\[2mm]  
  
 \end{tabular}

 \caption{Results of  fixing  a degraded handwritten line image. Errors made by the CRNN reading engine are shown in character level with the red color. R (GT): recognition by the CRNN \cite{flor2020CRNN} trained on clean images, R (D): recognition by the CRNN \cite{flor2020CRNN} trained on degraded images (better viewed in color),R (Generated): recognition by CRNN \cite{flor2020CRNN} trained on generated images (S2).}
 \label{fig:iam_recovery_1}
\end{center}
 \end{figure}

\begin{figure}[H]
  \begin{center}
 \begin{tabular}{l@{\hskip 0.1in}l}
    GT image: &\includegraphics[width=80mm, height=5mm]{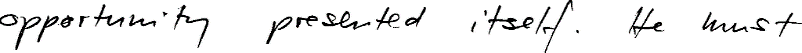}
  \\[2mm]  
   GT text: & \small opportunity presented itself . He must
  \\[2mm]  
   R(GT): & \small oportunity presented itsel\textcolor{red}{y}f . He must
  \\[2mm]  
 
   Distorted: & \includegraphics[width=80mm, height=5mm]{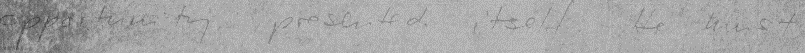}
  \\[2mm]  
   R(D): & \small  \textcolor{red}{o en a wao on}
  \\[2mm]  
   R(GT): & \small  \textcolor{red}{\#A}
  \\[2mm]  
   Baseline: &\includegraphics[width=80mm, height=5mm]{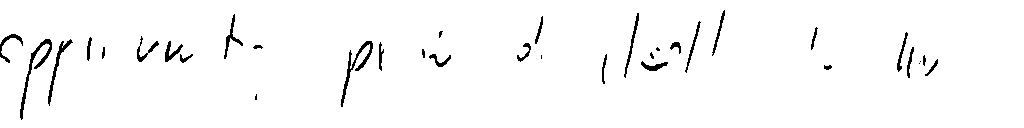}
  \\[2mm]  
 
 R(GT): & \small p\textcolor{red}{r ". Un} t. pr\textcolor{red}{i . 1lial! '- U.}
  \\[2mm]  
 
  cGAN \cite{souibgui2020conditional}: &\includegraphics[width=80mm, height=5mm]{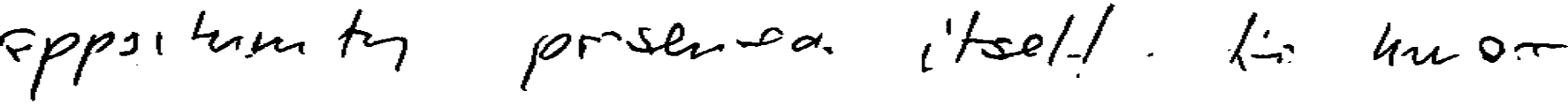}
  \\[2mm]  
   R(GT): & \small \textcolor{red}{rps hm to} prsen\textcolor{red}{ca} itsel\textcolor{red}{ ! }. \textcolor{red}{Lis unor}

  \\[2mm]  
  
   \textbf{Ours (S1)}: &\includegraphics[width=80mm, height=5mm]{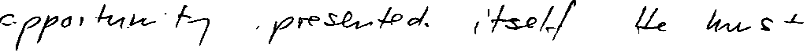}
  \\[2mm]  
   R(GT): & \small \textcolor{red}{s}portunty presented . itself He m\textcolor{red}{e}s\textcolor{red}{e}
  \\[2mm]  
    R(Generated): & \small aportun\textcolor{red}{-}ty presented itself . H\textcolor{red}{h}e must
  \\[2mm]  
  
  Ours (S2): &\includegraphics[width=80mm, height=5mm]{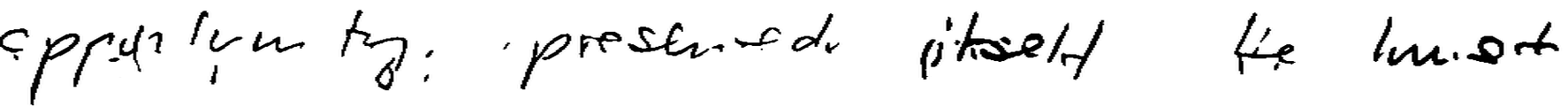}
  \\[2mm]  
   R(GT): & \small \textcolor{red}{s}p\textcolor{red}{en ' }in\textcolor{red}{-}ty \textcolor{red}{.} presen\textcolor{red}{-}ed\textcolor{red}{- ritsel} He \textcolor{red}{I}m\textcolor{red}{i}st
  \\[2mm]   
    R(Generated): & \small \textcolor{red}{s}p\textcolor{red}{an}t\textcolor{red}{i}n\textcolor{red}{-} ty presened \textcolor{red}{p}itsel\textcolor{red}{d} . \textcolor{red}{b}e \textcolor{red}{l}m\textcolor{red}{i}st .
  \\[2mm]

 \end{tabular}

 \caption{Results of  fixing  a highly degraded handwritten line image. Errors made by the CRNN reading engine are shown in character level with the red color. R (GT): recognition by  the CRNN \cite{flor2020CRNN} trained on clean images, R (D): recognition by the CRNN \cite{flor2020CRNN} trained on degraded images, R (Generated): recognition by CRNN \cite{flor2020CRNN} trained on generated images (S2).}
 \label{fig:iam_recovery_2}
\end{center}
 \end{figure}

\begin{figure}[H]
  \begin{center}
 \begin{tabular}{l@{\hskip 0.1in}l}
    GT image: &\includegraphics[width=80mm, height=5mm]{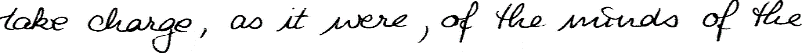}
  \\[2mm]  
   GT text: & \small take charge , as it were , of the minds of the
  \\[2mm]  
   R(GT): & \small take change , as it were , of the mi\textcolor{red}{s}nds of the
  \\[2mm]  
 
   Distorted: & \includegraphics[width=80mm, height=5mm]{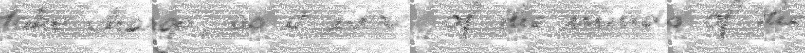}
  \\[2mm]  
   R(D): & \small  \textcolor{red}{Poe oroe wo h ore , }of the \textcolor{red}{onay} of the
  \\[2mm]  
   R(GT): & \small   \textcolor{red}{AHAH}
  \\[2mm]  
  Baseline: &\includegraphics[width=80mm, height=5mm]{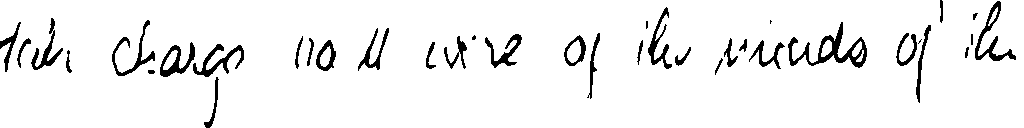}
  \\[2mm]  
   R(GT): & \small \textcolor{red}{Ari} charg\textcolor{red}{a} \textcolor{red}{n}a 4 e\textcolor{red}{iv}e of i\textcolor{red}{ke}s \textcolor{red}{n}i\textcolor{red}{s}nd\textcolor{red}{o} of \textcolor{red}{iln}
  \\[2mm]   

  cGAN \cite{souibgui2020conditional}: &\includegraphics[width=80mm, height=5mm]{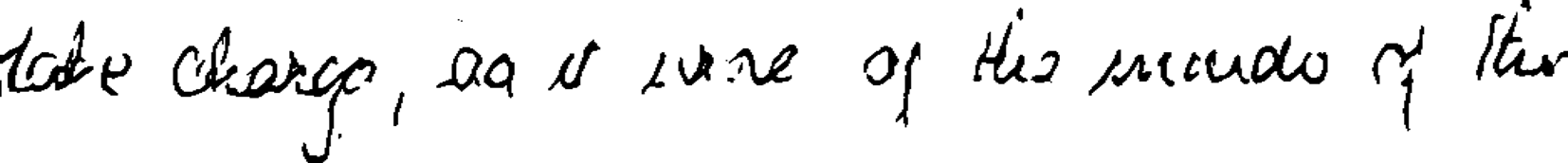}
  \\[2mm]  
   R(GT): & \small  \textcolor{red}{lod}e cha\textcolor{red}{n}g\textcolor{red}{a} , a \textcolor{red}{w} i\textcolor{red}{use} of this m\textcolor{red}{o}nd\textcolor{red}{o} of H\textcolor{red}{is}
  \\[2mm]  

  \textbf{Ours (S1)}: &\includegraphics[width=80mm, height=5mm]{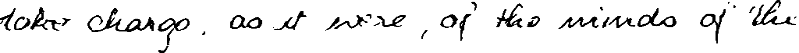}
  \\[2mm]  
   R(GT): & \small \textcolor{red}{Lo}t\textcolor{red}{r} chang\textcolor{red}{o} , a\textcolor{red}{o} \textcolor{red}{s}t w\textcolor{red}{is}e , \textcolor{red}{a}f th\textcolor{red}{o} minds \textcolor{red}{g "} the
  \\[2mm]  
    R(Generated): & \small  toke\textcolor{red}{r}cha\textcolor{red}{n}g\textcolor{red}{o} , as it \textcolor{red}{m}e\textcolor{red}{s}e , of the minds of the
  \\[2mm]  
  
   Ours (S2): &\includegraphics[width=80mm, height=5mm]{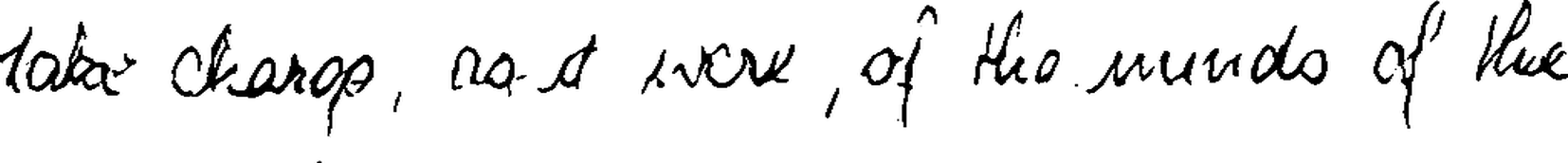}
  \\[2mm]  
   R(GT): & \small  \textcolor{red}{L}a\textcolor{red}{t} cha\textcolor{red}{nop} , \textcolor{red}{n}a \textcolor{red}{d} w\textcolor{red}{o}re , of th\textcolor{red}{o} \textcolor{red}{nu}nd\textcolor{red}{o} of the
  \\[2mm]   
   R(Generated): & \small  \textcolor{red}{l}ake cha\textcolor{red}{nop} , \textcolor{red}{h}a\textcolor{red}{t} w\textcolor{red}{o}re , of the m\textcolor{red}{u}nd\textcolor{red}{o} of the
  \\[2mm]

 \end{tabular}

 \caption{Results of  fixing  an extremely degraded handwritten line image. Errors made by the CRNN reading engine are shown in character level with the red color. R (GT): recognition by CRNN \cite{flor2020CRNN} trained on clean images, R (D): recognition by the CRNN \cite{flor2020CRNN} trained on degraded images, R (Generated): recognition by the CRNN \cite{flor2020CRNN} trained on generated images (S2).}
 \label{fig:iam_recovery_3}
\end{center}
 \end{figure}

Furthermore, we show some qualitative results in Figures \ref{fig:iam_recovery_1}, \ref{fig:iam_recovery_2} and \ref{fig:iam_recovery_3}, to visualize the performances of the different methods. Of course, reading the degraded image by a model trained on the GT clean images is not a suitable option. Also, training a model on degraded images is not improving the recognition, especially in the hard scenarios. That is why, enhancing the image and then reading it  is the better solution. As it can be seen, our  method is better in this practice especially than  the baseline cGAN (without a recognizer), because ours  is a text conservative method. Hence, it maps the image to a clean but readable domain, while the basic GAN is mapping the image to a visually clean version, without taking the text into consideration (see Figures \ref{fig:iam_recovery_2} and \ref{fig:iam_recovery_3}).

{For the sake of  more confirmation and to prove that our model is independent from the used recognizer,  we took a different  state-of-the-art HTR that is  Puigcerver's model~\cite{Puigcerver2017} trained on the GT images of IAM and KHATT original datasets.  Then, we carried a binarization stage to our degraded databases using different methods (including ours) and measure the final recognition performance.  As it can be seen 
from Table~\ref{tab_binarization_cer_perf}, our proposed binarization method enhances the performance of the recognizer compared to the use of images binarized by the classic methods \cite{otsu1979threshold,sauvola2000adaptive} or the recent cGAN's based one. Also, we can confirm the efficiency of our proposed binarization method compared to the baseline cGAN which did not integrate the text readability information.
}

\FloatBarrier
\begin{table}[!h]
\caption{Impact of the proposed binarization method (scenario S1) on the recognition performance  by a HTR system.}\label{tab_binarization_cer_perf}
\centering
\small
\begin{tabular}{|p{3.5cm}|p{4cm}|p{1.5cm}|p{1.cm}|}
\noalign{\smallskip}
\hline  
Dataset & Binarization Method &  CER\%  & WER\%          \\ \hline
\multirow{5}{*}{degraded-KHATT} & Otsu \cite{otsu1979threshold}  &   54.28 &  85.42       \\ \cline{2-4}
& Sauvola \cite{sauvola2000adaptive}&  58.42  &   99.57       \\ \cline{2-4}
& cGAN \cite{souibgui2020conditional} & 28.32 &  53.96 \\ \cline{2-4}
& Baseline cGAN  &     28.61      &    53.73     \\ \cline{2-4} 
& \textbf{Ours (S1)} & \textbf{26.57} & \textbf{52.31}        \\ \hline
\multirow{5}{*}{degraded-IAM}&Otsu \cite{otsu1979threshold} &   62.62 &    81.96      \\ \cline{2-4}
&Sauvola \cite{sauvola2000adaptive} & 72.48  &  98.00      \\ \cline{2-4}
&cGAN \cite{souibgui2020conditional}&  27.21 & 58.18\\ \cline{2-4}
&Baseline cGAN  &      31.31   &  61.79      \\ \cline{2-4} 
&\textbf{Ours (S1)}   & \textbf{25.79} & \textbf{56.43}       \\ \hline
\end{tabular}
\end{table}

\subsubsection{H-DIBCO Competitions}

\begin{table}[!b]
\centering
\small
\caption{{Comparative results of our proposed method on \textit{H-DIBCO 2012} Dataset for document binarization. Avg $=$ (PSNR $+$ FM $+$ Fps $+$ (100 $-$ DRD)) $/$ 4.}}
\label{tab:dibco_12_results}
\begin{tabular}{|l|l|l|l|l|l|}
\noalign{\smallskip}
\hline
Method                           & PSNR  & FM & Fps   & DRD    & Avg \\ \hline

Otsu \cite{otsu1979threshold} & 15.03 &80.18& 82.65& 26.46&62.85 \\ \hline
Sauvola et al. \cite{sauvola2000adaptive}& 16.71&82.89& 87.95 & 6.59& 70.24\\ \hline
Guo et al. \cite{Guo2019}& 17.86 &86.40& 89.00 &4.67& 72.14 \\ \hline
Zhao et al. \cite{zhao2019document} & 21.91&94.96 &96.15 & 1.55& 77.86 \\ \hline
Competition winner \cite{Pratikakis2012ICFHR} & 21.80&89.47 &90.18 & 3.44& 74.50\\ \hline
Kang et al. \cite{kang2021complex}&21.37&95.16& \textbf{96.44} & \textbf{1.13}&\textbf{77.96} \\ \hline
\textbf{Ours (S1)}  &  {16.29} &  {79.25} &  {85.96} & {7.33} &  {68.54}  \\ \hline
\textbf{Ours (S1) + Fine-tuning}    &  \textbf{22.00} &  \textbf{95.18} &  {94.63} & {1.62} &  {77.54}  \\ \hline
\end{tabular}
\end{table}

After demonstrating the suitability of our proposed method in recovering clean and readable images from  highly degraded ones. In what follows, we validate it in H-DIBCO competition on handwritten document binarization,  {using H-DIBCO 2012 \cite{Pratikakis2012ICFHR}, H-DIBCO 2016, DIBCO 2017 \cite{Pratikakis2017} and H-DIBCO 2018 \cite{Pratikakis2018icfhr}}.  Since our model was designed to enhance line images with a size of $128 \times 1024$, we binarize H-DIBCO images in form of patches having the same dimensions.   We compare with the recent state of the art approaches, the winners of the different competitions \cite{Pratikakis2012ICFHR} and the classic binarization methods \cite{otsu1979threshold,sauvola2000adaptive}.   To clean H-DIBCO images, and since they are formed of Latin script text, we used our pretrained model on the developed degraded-IAM dataset. Two scenarios were investigated: Using the model directly to clean the images, or fine tuning it with a similar distribution before using it. For fine tuning, we used the other DIBCO and H-DIBCO versions \cite{Pratikakis2018icfhr} and the Palm-Leaf dataset \cite{burie2016icfhr2016}.  It is to note also that since the DIBCO datasets are not holding the text information, we removed the recognizer component during fine tuning process, we have frozen also the batch normalisation layers of the generator and we trained the architecture for one only epoch to keep the learned knowledge of the degraded-IAM.  During cleaning, we feed our model  with the original degraded image in two forms: A normal condition and  a vertically flipped version. Thus, we produce two instances of the recovered  images.  The flipped image is, then, re-flipped again to the normal condition. After that, a voting method is used to produce the final binarized image, by assigning zero to the pixel value (black) if it is indeed black in the two produced images by our model. We found that this led to a better result instead of using just one image condition. 

{We start our experiments with H-DIBCO 2012, the obtained results are given in Table~\ref{tab:dibco_12_results}. As it can be seen, our model leads to competitive results to the state of the art approaches, with a superiority in two metrics (PSNR and FM). However, we can see that the model proposed in \cite{kang2021complex} is better in term of $F_{ps}$ and DRD. 
Then, we tested our method on a more recent dataset which is H-DIBCO 2016 \cite{Pratikakis2016}. As presented in Table~\ref{tab:dibco_16_results}, our model gives the state of the art compared to all the methods in the three metrics PSNR, FM and DRD and in the overall average. }

\begin{table}[h]
\centering
\small
\caption{{Comparative results of our proposed method on \textit{H-DIBCO 2016} Dataset for document binarization. Avg $=$ (PSNR $+$ FM $+$ Fps $+$ (100 $-$ DRD)) $/$ 4.}}
\label{tab:dibco_16_results}
\begin{tabular}{|l|l|l|l|l|l|}
\noalign{\smallskip}
\hline
Method                           & PSNR  & FM & Fps   & DRD    & Avg \\ \hline
Otsu \cite{otsu1979threshold} & 17.80 & 86.61& 88.67 & 7.46&71.40 \\ \hline
Sauvola et al. \cite{sauvola2000adaptive}& 16.42&82.52& 86.85&  5.56&70.05\\ \hline
Vo et al. \cite{Vo2018} & 19.01 &90.10& 93.57& 3.58&74.77 \\ \hline
Guo et al. \cite{Guo2019}& 18.42&88.51& 90.46&  4.13&73.31 \\ \hline
He and Schomaker \cite{He2019} &19.60& 91.40& 94.30& 2.90&75.6\\ \hline
Zhao et al. \cite{zhao2019document} & 19.64&91.66 &\textbf{94.58} &2.82&75.76 \\ \hline
Competition winner \cite{Pratikakis2016} &18.11&87.61& 91.28 & 5.21&72.94 \\ \hline
Bera et al.  \cite{Bera2021} &18.94 &90.43 &91.66 &3.51&74.38 \\ \hline
Kang et al. \cite{kang2021complex}&  19.18 &93.09& 94.85&3.03&76.02 \\ \hline
\textbf{Ours (S1)}  &  {14.26} &  {69.52} &  {78.01} &  {12.11} & {62.42}  \\ \hline
\textbf{Ours (S1) + Fine-tuning}    &  \textbf{21.85} &  \textbf{94.95} &  {94.55} &  \textbf{1.56}   & \textbf{77.44} \\ \hline
\end{tabular}
\end{table}

 {Next,   we tested with DIBCO 2017, which contains a mix of handwritten and printed degraded documents images. The results in Table~\ref{tab:dibco_17_results} show that our model is not superior in this dataset, but it is competitive with the best approaches. We note that our model performance was affected by the type of binarized documents in this dataset, which contain several printed documents, while our model is designed essentially for  the handwritten text.  Finally, we tested with the most recent H-DIBCO 2018~\cite{Pratikakis2018icfhr}. The results are shown in  Table~\ref{tab:dibco18_results}, where we compare with the most recent state of the art results, the winner of the H-DIBCO 2018 competition and the classic binarization methods. The performance of our model is superior than the different approaches in term of PSNR, FM, $F_{ps}$ and average score.
 }
 
{Out of the obtained results in the different datasets, we can say that the classic thresholding methods \cite{otsu1979threshold,sauvola2000adaptive} have a moderate performance compared to the recent deep learning approaches. Also, we can notice that if our model using only degraded-IAM for training,  does not reach a satisfactory result because there is a domain gap between the training and testing data. However, fine tuning our model with the similar datasets leads to the best performance compared to all the state of the art methods in H-DIBCO 2016 and H-DIBCO 2018. While having a competitive result with the best approach in H-DIBCO 2012 that is  \cite{kang2021complex}, where we obtain a superior PSNR and FM scores.  We can conclude also that our model is more suitable for binarizing the handwritten images, since it was  pre-trained on the developed degraded-IAM dataset before the fine tuning stage.}


\begin{table}[h]
\centering
\small
\caption{{Comparative results of our proposed method on \textit{DIBCO 2017} Dataset for document binarization. Avg $=$ (PSNR $+$ FM $+$ Fps $+$ (100 $-$ DRD)) $/$ 4.}}
\label{tab:dibco_17_results}
\begin{tabular}{|l|l|l|l|l|l|}
\noalign{\smallskip}
\hline
Method                           & PSNR  & FM & Fps   & DRD    & Avg \\ \hline
Otsu \cite{otsu1979threshold} &13.85& 77.73& 77.89 & 15.54&  63.48  \\ \hline
Sauvola et al. \cite{sauvola2000adaptive}&14.25& 77.11& 84.1 & 8.85&  66.65 \\ \hline
Zhao et al. \cite{zhao2019document} &17.83&90.73& 92.58 & 3.58& 74.39  \\ \hline
Competition winner \cite{Pratikakis2017} & \textbf{18.28}&91.04& 92.86&  3.40& \textbf{74.69} \\ \hline
Kang et al. \cite{kang2021complex}& 15.85&\textbf{91.57}& \textbf{93.55} & \textbf{2.92}&  74.51  \\ \hline
Bera et al. \cite{Bera2021}  &15.45 &83.38 &89.43 &6.71&  70.38  \\ \hline
\textbf{Ours (S1)}  &  {13.54} &  {71.13} &  {80.39}  &  {9.60} & {63.86}   \\ \hline
\textbf{Ours (S1) + Fine-tuning}    &  {17.45} &  {89.8} &  {89.95} &  {4.03} &  {73.29}  \\ \hline
\end{tabular}
\end{table}

\begin{table}[h]
\centering
\small
\caption{Results for all methods on \textit{H-DIBCO 2018} Dataset for handwritten document binarization. Avg $=$ (PSNR $+$ FM $+$ Fps $+$ (100 $-$ DRD)) $/$ 4.}
\label{tab:dibco18_results}
\begin{tabular}{|l|l|l|l|l|l|}
\noalign{\smallskip}
\hline
Method                           & PSNR  & FM & Fps   & DRD    & Avg \\ \hline
Otsu \cite{otsu1979threshold}                    & 9.74  & 51.45     & 53.05 & 59.07   & 38.79 \\ \hline
Sauvola et al. \cite{sauvola2000adaptive}  & 13.78 & 67.81     & 74.08 & 17.69 & 59.50 \\ \hline
Adak et al. \cite{Pratikakis2018icfhr}             & 14.62 & 73.45     & 75.94 & 26.24   & 59.44\\ \hline
Souibgui et al. \cite{souibgui2020gan}               & 16.16 & 77.59     & 85.74 & 7.93  &  67.89  \\ \hline
Tamrin et al. \cite{tamrin2021}          & 17.04 & 83.08     & 88.46 & 5.09  &  70.87\\ \hline
Zhao et al. \cite{zhao2019document}             & 18.37 & 87.73     & 90.6  & 4.58 & 	73.03  \\ \hline
Competition winner \cite{Pratikakis2018icfhr}  & 19.11 & 88.34     & 90.24 & 4.92  & 	73.19 \\ \hline
Akbari et al. \cite{akbari2020binarization} & 19.17 & 89.05 & 93.65 & 4.80 & 74.26\\ \hline
Kang et al. \cite{kang2021complex}            & 19.39 & 89.71                         & 91.62                         &  \textbf{2.51} & 	74.55 \\ \hline
Dang et al. \cite{Dang2021Binarization}            & 19.81 & 91.26                         & 93.97                        &  3.42 & 	75.40 \\ \hline
Bera et al. \cite{Bera2021}    & 15.31&        76.84& 83.58& 9.58 & 66.53 \\ \hline
\textbf{Ours (S1)}                    & 13.88 & 65.06     & 73.46 & 12.86  & 	59.89 \\ \hline
 \textbf{Ours (S1) + Fine-tuning}                       &  \textbf{20.18}                          &  \textbf{92.41} &  \textbf{94.35} & 2.60      &  \textbf{76.08}                   \\ \hline
\end{tabular}
\end{table}


  \begin{figure*}[!h]
  \begin{center}
  
 \begin{tabular}{ccc}
 
    \includegraphics[width=0.30 \columnwidth, height=20mm]{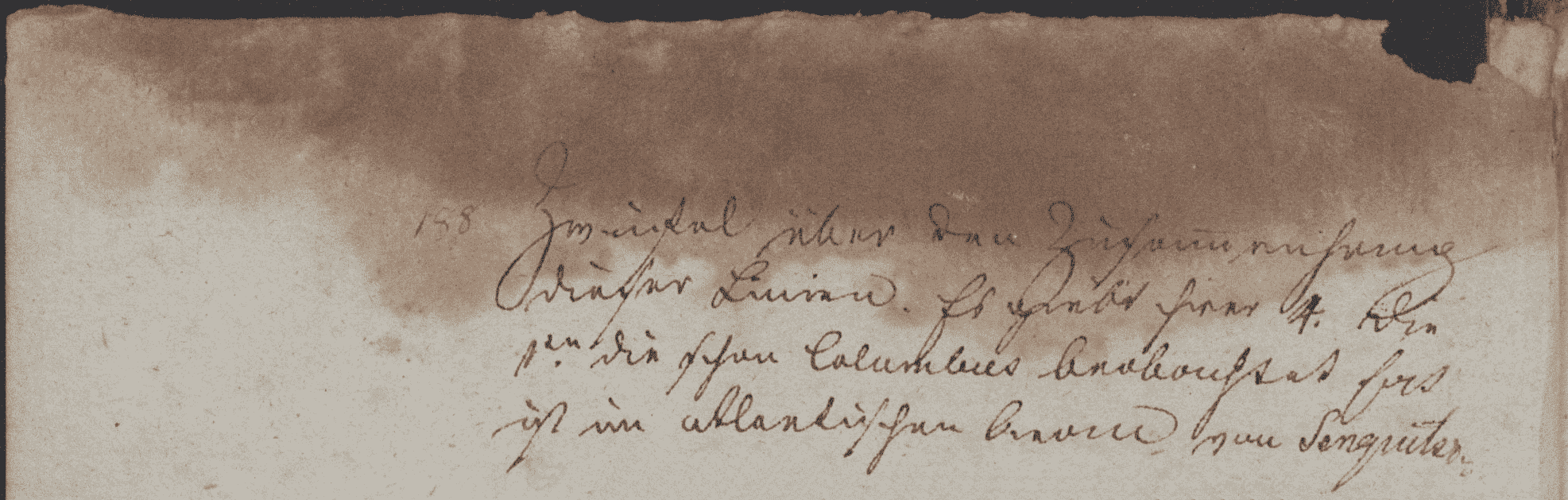} &
    \includegraphics[width=0.30 \columnwidth, height=20mm]{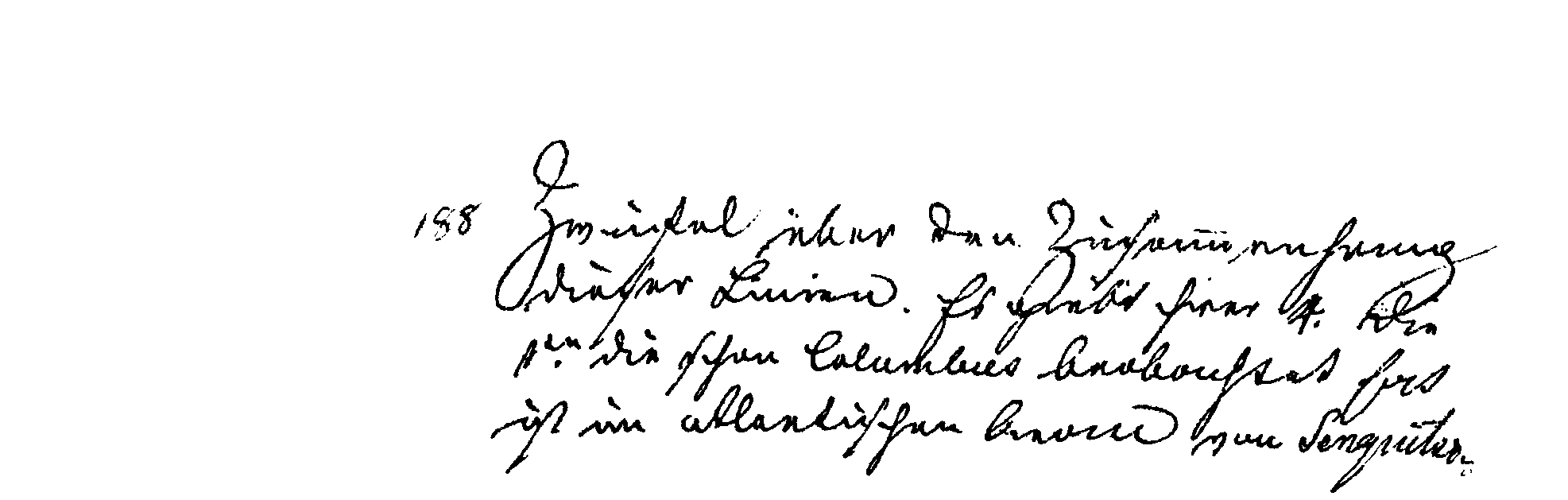}&
      \includegraphics[width=0.30 \columnwidth, height=20mm]{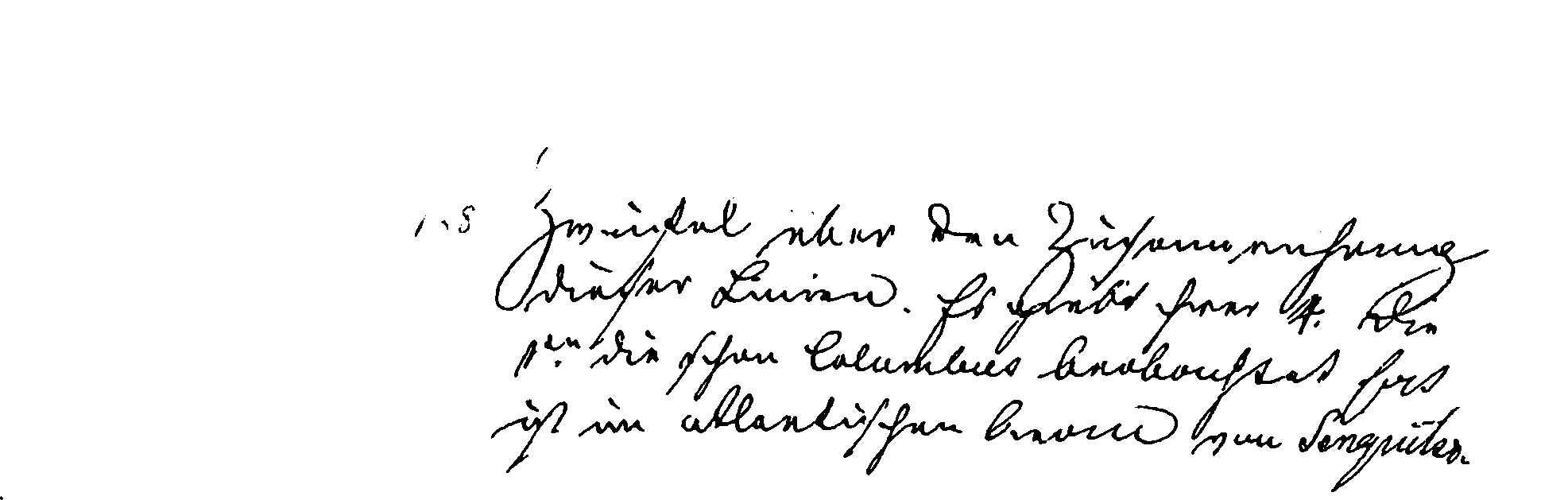}
    \\ \noalign{\smallskip} 
    
            \includegraphics[width=0.30 \columnwidth, height=20mm]{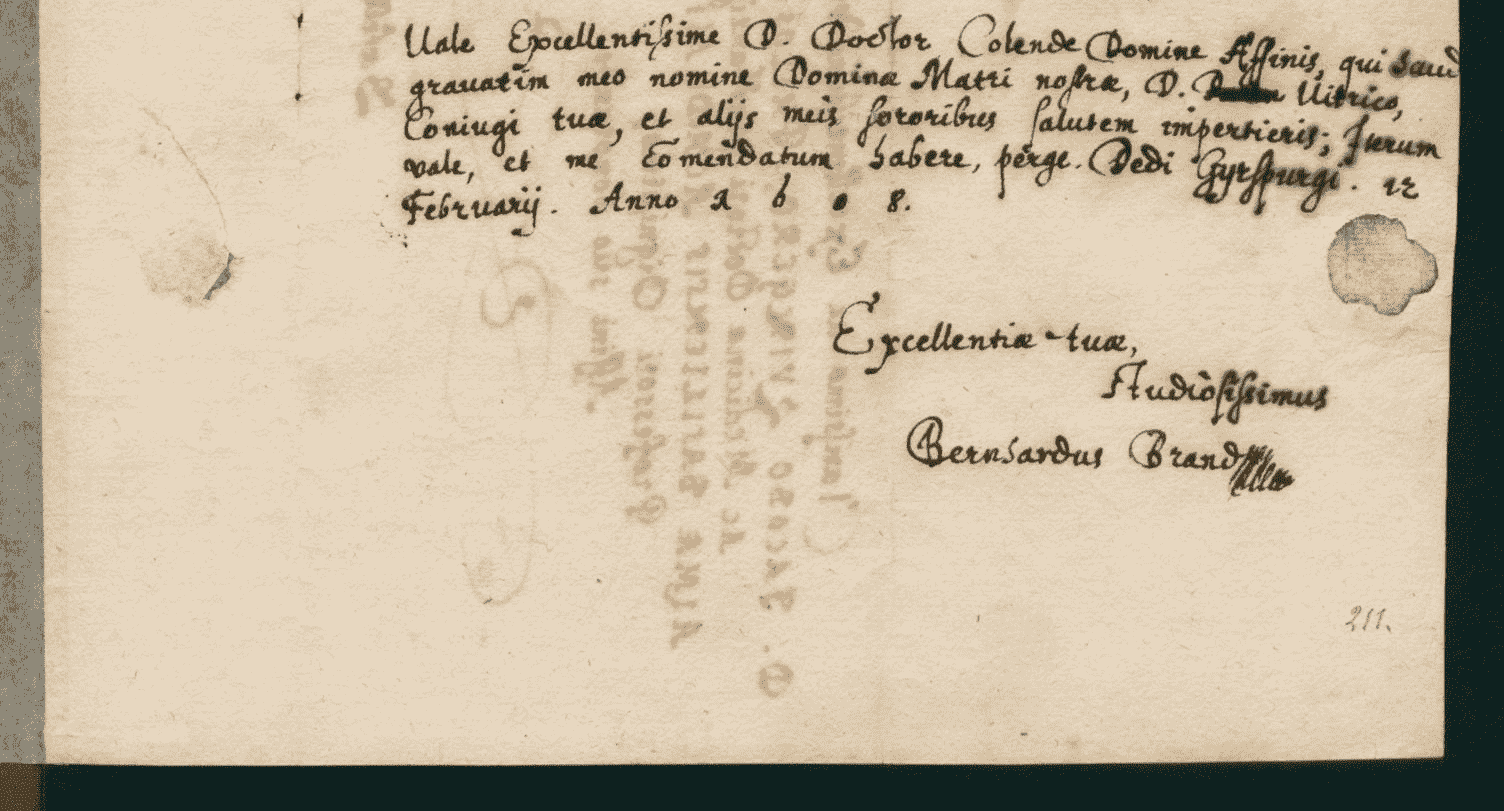} &
    \includegraphics[width=0.30 \columnwidth, height=20mm]{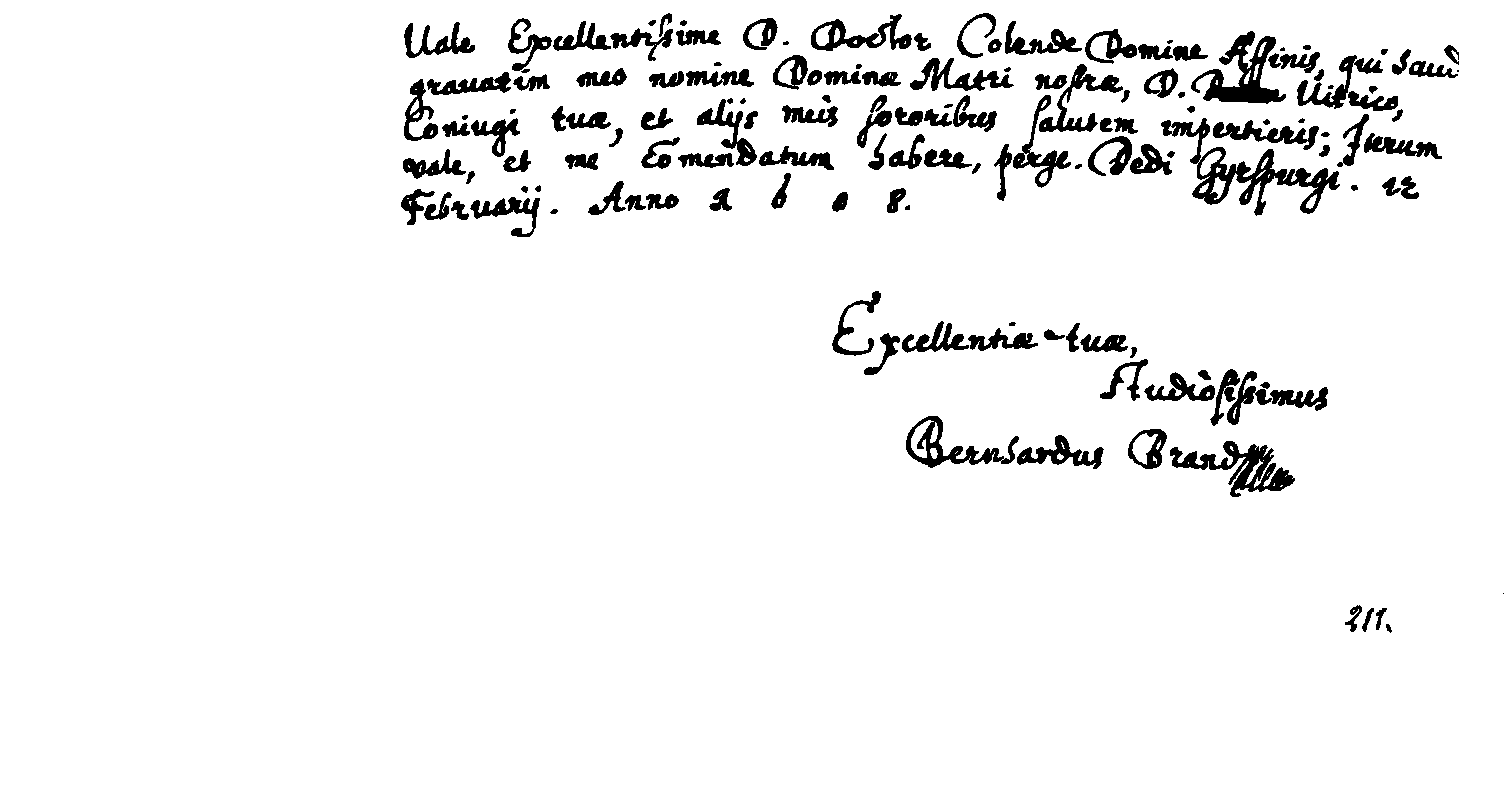}&
      \includegraphics[width=0.30 \columnwidth, height=20mm]{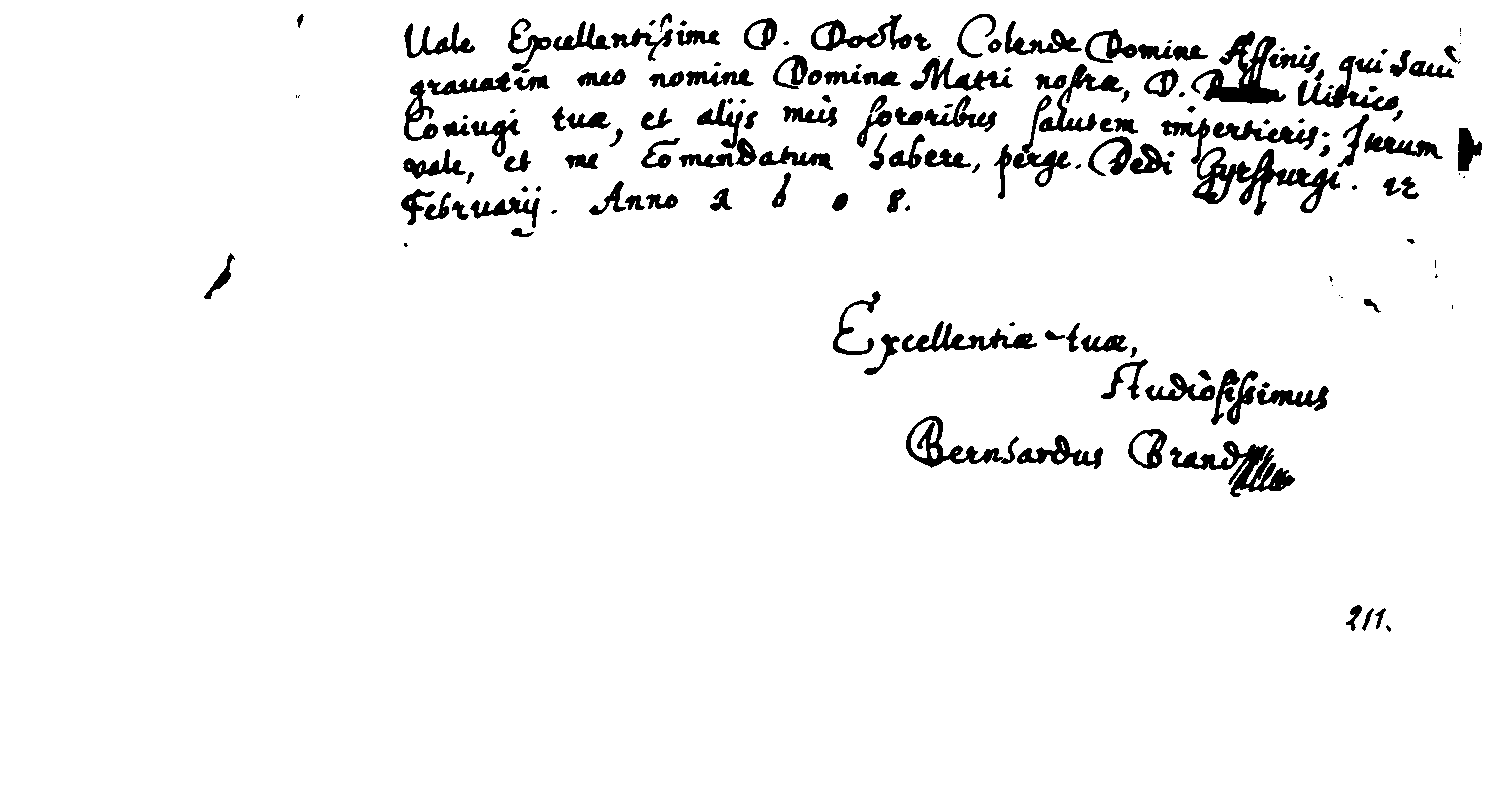}
    \\ \noalign{\smallskip}

     \includegraphics[width=0.30 \columnwidth, height=20mm]{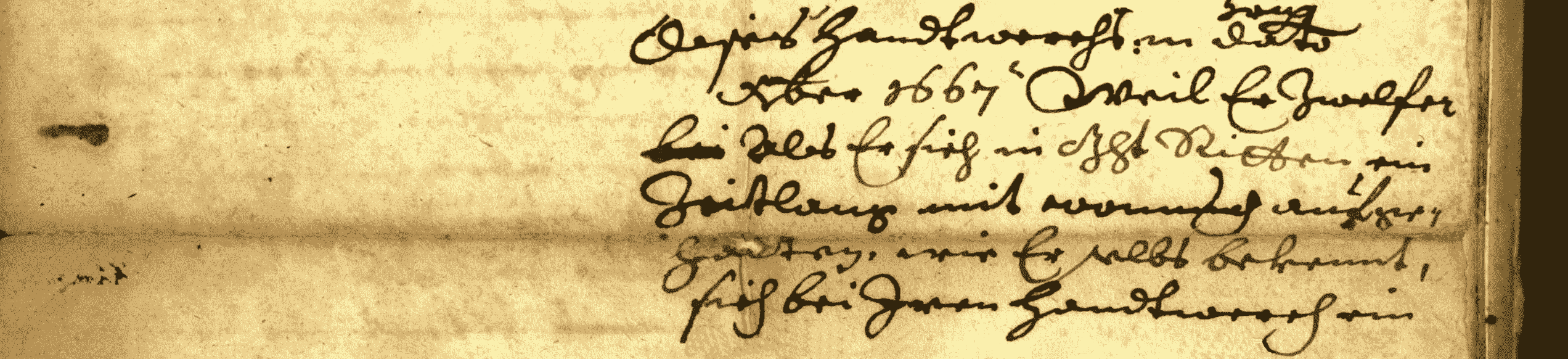} &
    \includegraphics[width=0.30 \columnwidth, height=20mm]{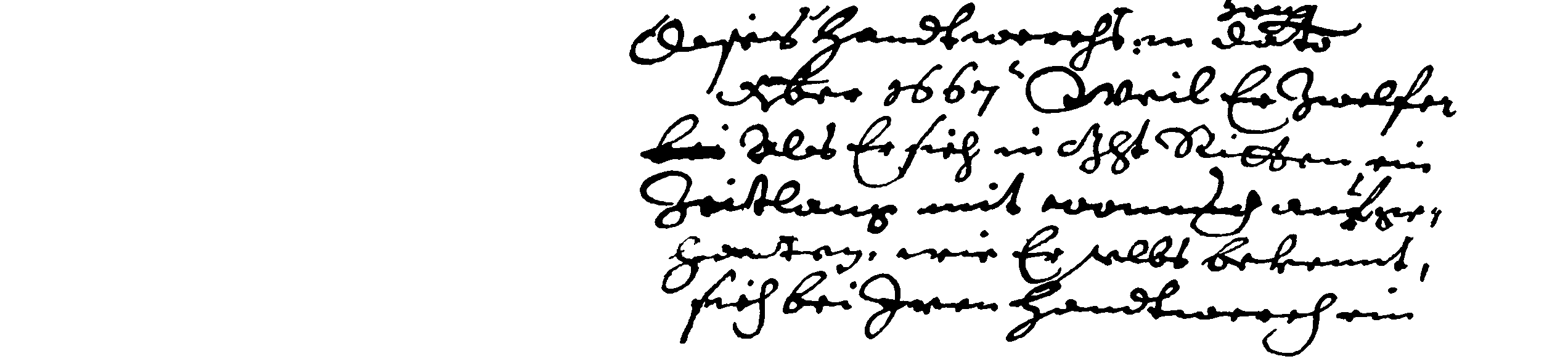}&
      \includegraphics[width=0.30 \columnwidth, height=20mm]{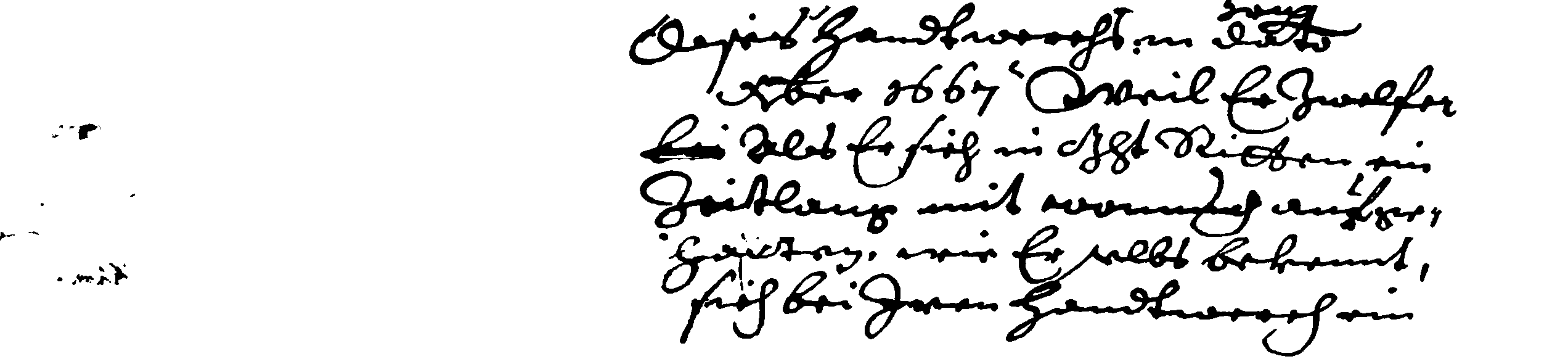}
    \\ \noalign{\smallskip} 
    
           
        \includegraphics[width=0.30 \columnwidth, height=20mm]{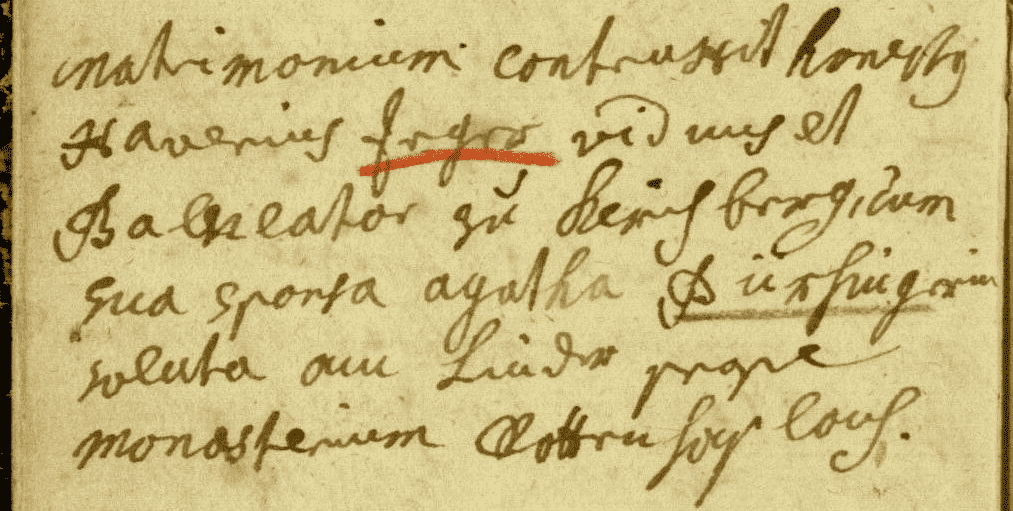} &
    \includegraphics[width=0.30 \columnwidth, height=20mm]{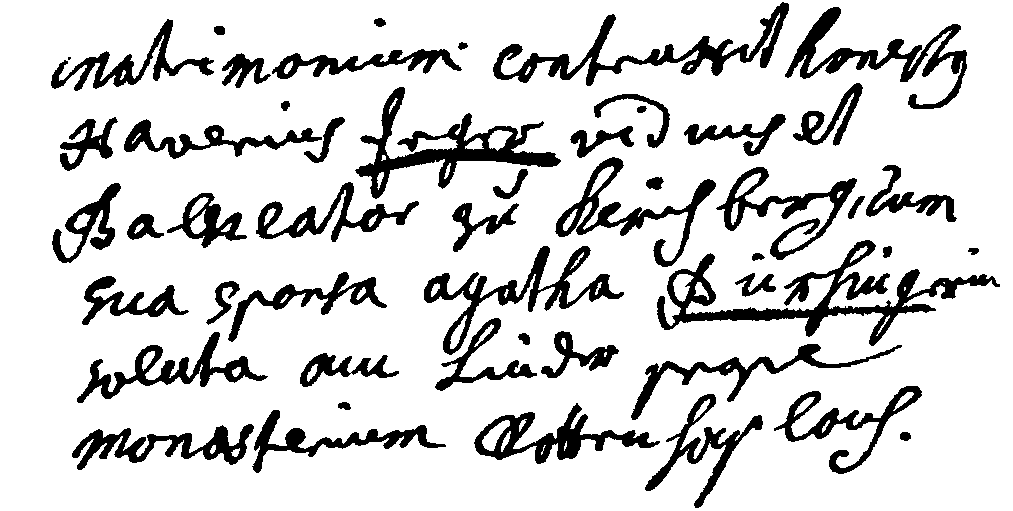}&
      \includegraphics[width=0.30 \columnwidth, height=20mm]{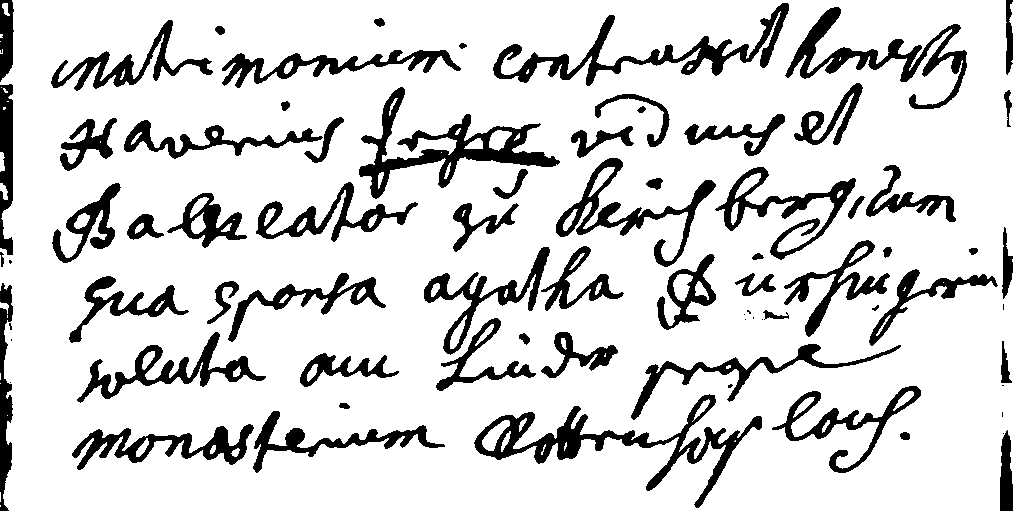}
    \\ \noalign{\smallskip}

 \end{tabular}

 \caption{Results of our  method in binarization of some samples from the H-DIBCO 2018 dataset. Images in columns are: Left: original image, Middle: GT image, Right: Binarized image using our proposed method.}
 \label{fig:results_dibco18_details}
\end{center}
 \end{figure*}


  \begin{figure}[!h]
  \begin{center}
   \begin{tabular}{cc}
 
    \includegraphics[width=0.4 \columnwidth, height=25mm]{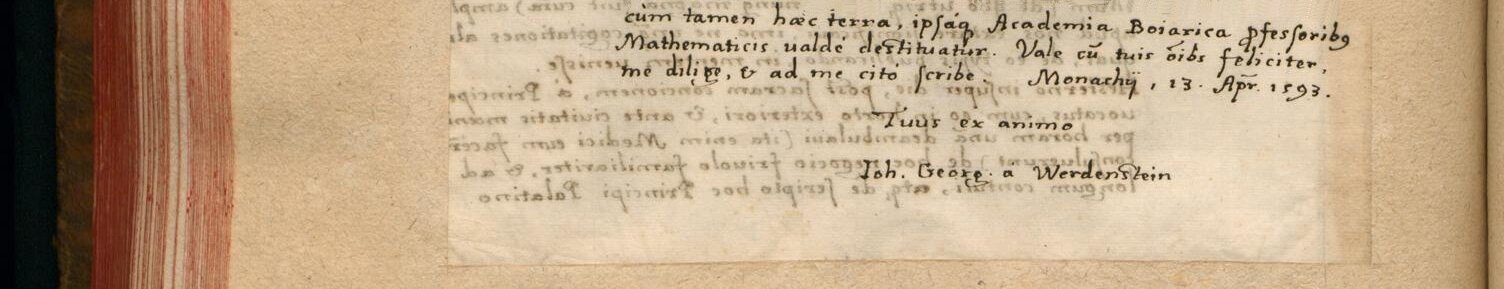} &
    \includegraphics[width=0.4 \columnwidth, height=25mm]{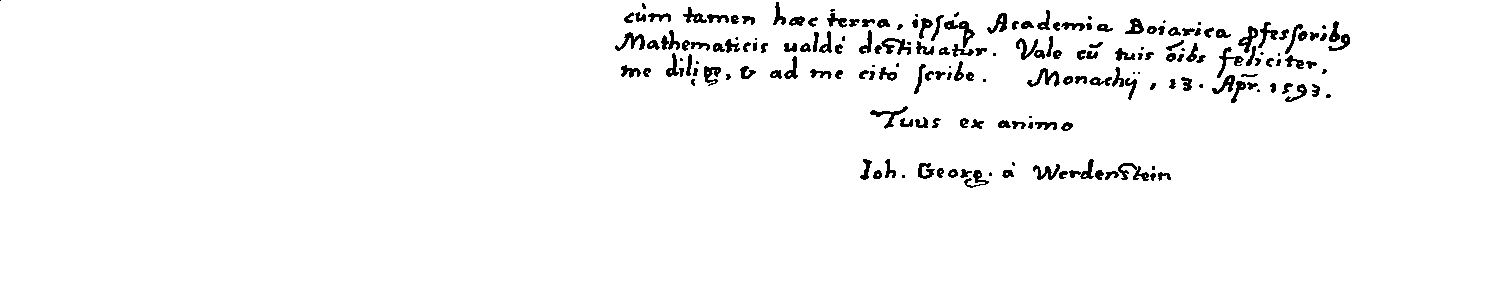}\\ \noalign{\smallskip} 
    Original  & GT 
    \\ \noalign{\smallskip}     
    \includegraphics[width=0.4 \columnwidth, height=25mm]{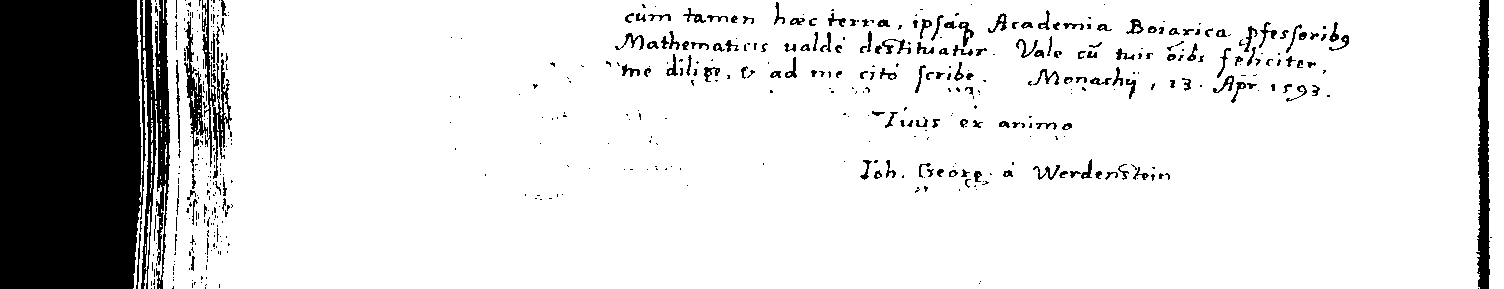}&
    \includegraphics[width=0.4 \columnwidth, height=25mm]{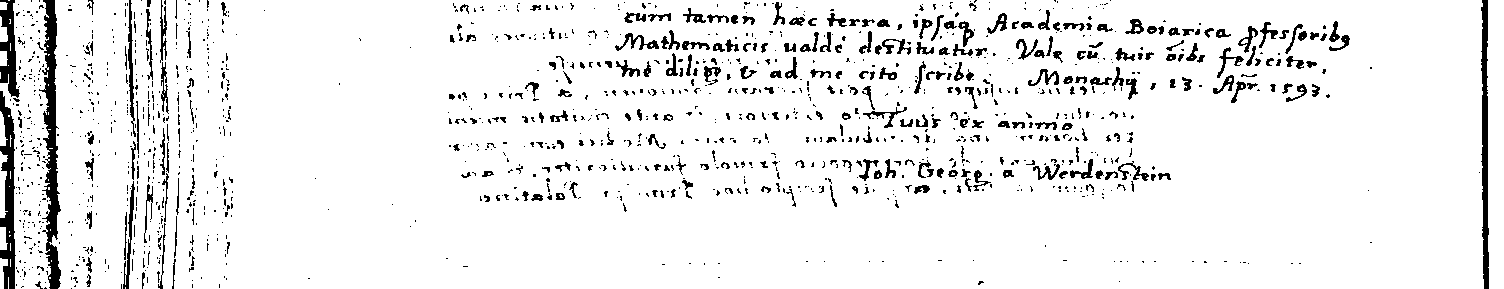} 
    \\\noalign{\smallskip}   
    Otsu \cite{otsu1979threshold} & Sauvola et al. \cite{sauvola2000adaptive}
    \\\noalign{\smallskip}   
    
    \includegraphics[width=0.4 \columnwidth, height=25mm]{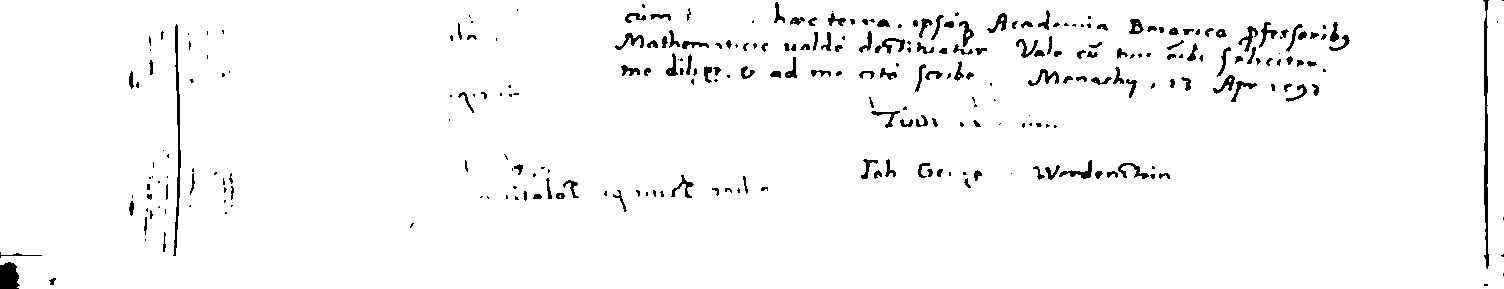}&
    \includegraphics[width=0.4 \columnwidth, height=25mm]{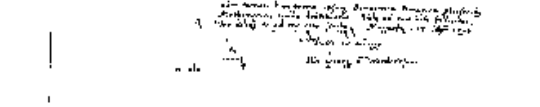} 
    \\\noalign{\smallskip}   
    Souibgui et al. \cite{souibgui2020gan} & H-DIBCO 2018 Winner \cite{Pratikakis2018icfhr} 
    \\\noalign{\smallskip}

     \includegraphics[width=0.4 \columnwidth, height=25mm]{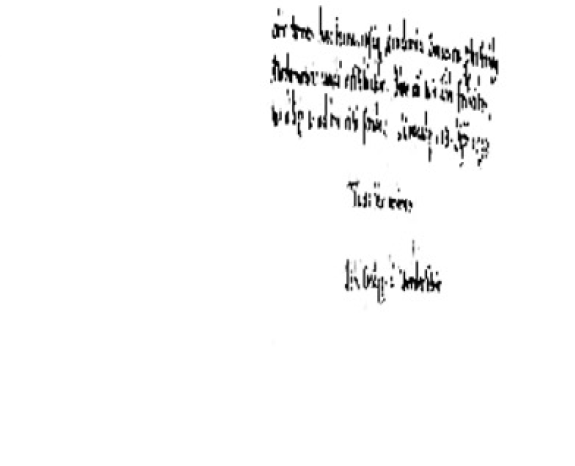}&
    \includegraphics[width=0.4 \columnwidth, height=25mm]{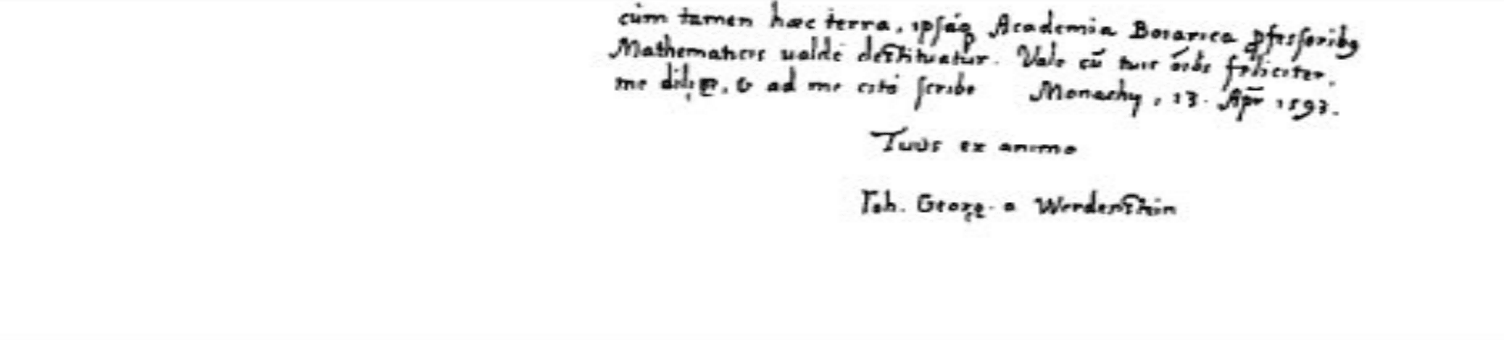} 
    \\\noalign{\smallskip}   
    Kang et al. \cite{kang2021complex} & Dang et al. \cite{Dang2021Binarization}   
    \\\noalign{\smallskip}

    \includegraphics[width=0.4 \columnwidth, height=25mm]{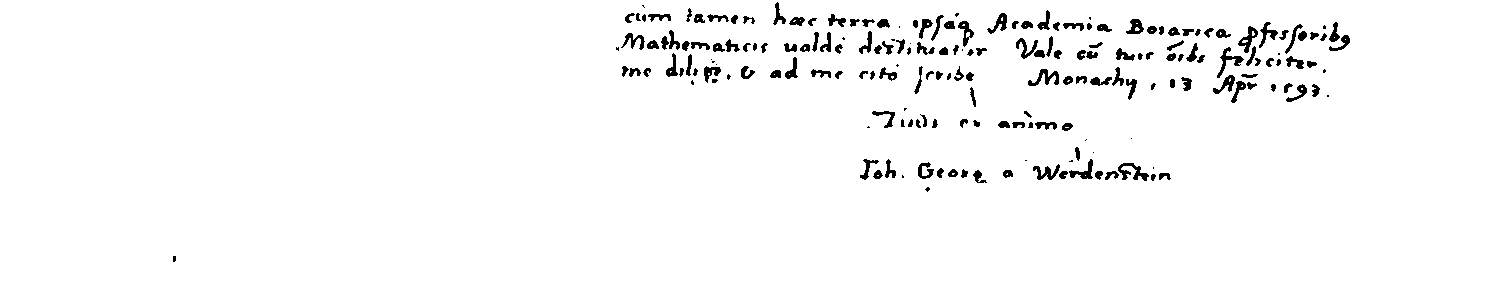} 
    \\\noalign{\smallskip}   
     Ours  
    \\\noalign{\smallskip}

 \end{tabular}

 \caption{Results of the  different enhancement  on the  sample 4, from H-DIBCO 2018 Dataset.}
 \label{fig:results_dibco18}
\end{center}
 \end{figure}

\begin{figure}[!h]
  \begin{center}
  
 \begin{tabular}{c}
 
    \includegraphics[width=0.9\columnwidth, height=15mm]{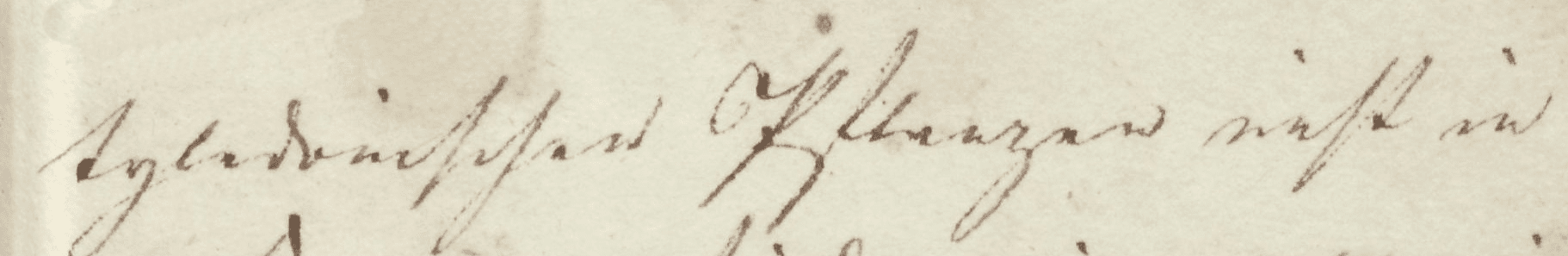} \\
      \noalign{\smallskip} 
    Original   
     \\ \noalign{\smallskip}   
    
    \includegraphics[width=0.9\columnwidth, height=15mm]{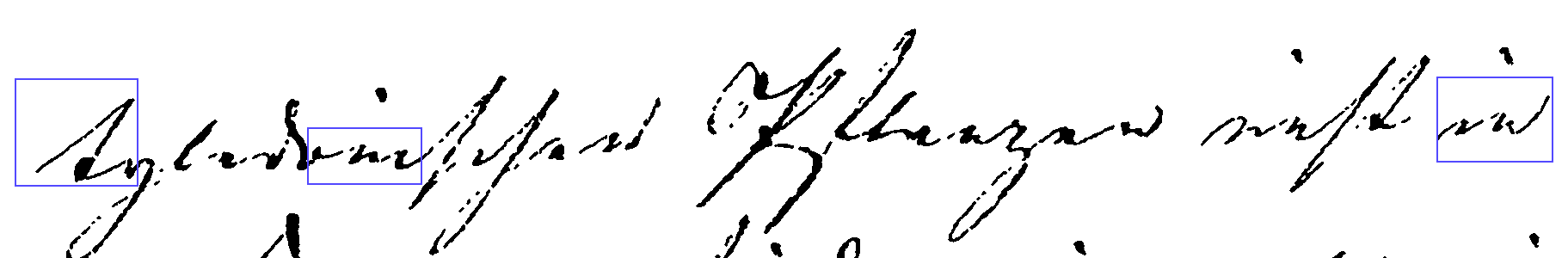} \\
      \noalign{\smallskip} 
    GT   
     \\ \noalign{\smallskip}    
 
     \includegraphics[width=0.9\columnwidth, height=15mm]{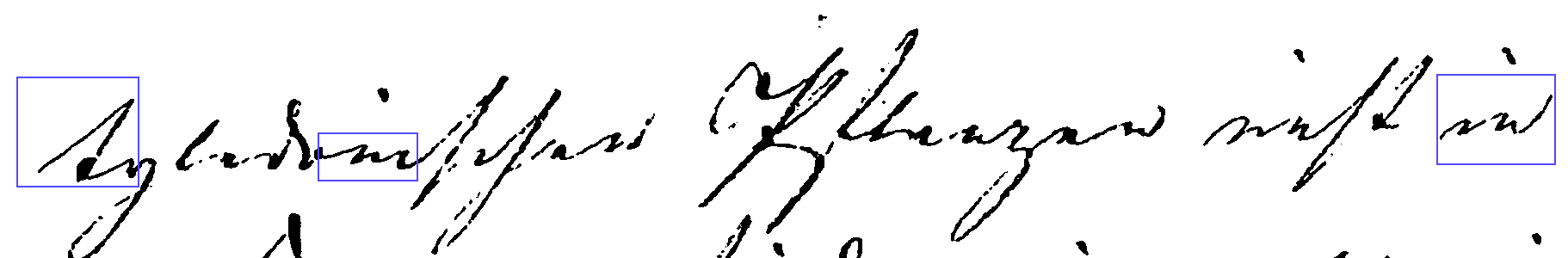} \\ 
      \noalign{\smallskip} 
    Enhanced (Ours)   
     \\ \noalign{\smallskip}   
   
 \end{tabular}

 \caption{Qualitative results of our proposed method evaluated on an a part taken from a sample from H-DIBCO 2018 Dataset (Pixels restoration).}
 \label{fig:results_dibco18Restoration}
\end{center}
 \end{figure}

Finally, we show some qualitative results about the binarization performance in Figure~\ref{fig:results_dibco18} that demonstrates our method superiority compared to the other ones in this task. Also, we provide the binarization result of other images from the H-DIBCO 2018 dataset in Figure ~\ref{fig:results_dibco18_details} where we obtained images that are very close to the GT. Moreover, Figure~\ref{fig:results_dibco18Restoration} shows an example where our method can even complete the missing pixels (that are not existent in the GT image), to provide a more readable text.

\FloatBarrier
\FloatBarrier

\subsubsection{Dataset selection for the fine-tuning stage}

{ Selecting the right dataset for fine-tuning will improve the binarization performance. Thus, in this section, we study the impact of the different datasets on the binarization process carried out on the H-DIBCO 2016. For all our experiments, we tested using the other variations of DIBCO and H-DIBCO (from 2009 to 2018), except DIBCO 2019 since it has different distributions in term of degradation and document types. Also, we tested including similar datasets which were developed for binarization task, namely,  Palm-Leaf \cite{burie2016icfhr2016},   Nabuco \cite{Nabuco2011}  and  Bickley-diary \cite{Bickley1926}. Images contained in these datasets suffer from different kinds of degradation, such as water stains, ink bleed-through, and significant foreground text intensity.  As it can be seen from Table~\ref{tab:dibco_ablation_results16}, using our model trained only on degraded-IAM  leads to poor results, thus, a fine-tuning stage is necessary. Using the H-DIBCO images for fine-tuning  improves the performance with a slight superiority than using the DIBCO ones. This can be explained by the type of text, because our model is pretrained to binarize the handwritten text. However, using H-DIBCO and DIBCO at the same time is a better option. Because, DIBCO contains a useful types of degradation that can be learned to be cleaned by our model even with a printed text. Also, adding other datasets is sometimes useful, but in other times deteriorates the performance. This can be noticed when adding the Palm-Leaf dataset which improves the binarization, while adding the Nabuco or the Bickley-diary is leading the model to learn a non suitable parameters for H-DIBCO 2016. This  can be justified by the similar domain (degradation and text types) between H-DIBCO 2016 and Palm-Leaf distributions, while it is different with the other datasets.}

\FloatBarrier
\begin{table}[h!]
\centering
\caption{Impact of the fine-tuning data selection on the binarization performance on H-DIBCO \textbf{2016} Dataset.}
\label{tab:dibco_ablation_results16}
\begin{tabular}{|c|c|c|c|c|c|c|}
\noalign{\smallskip}
\hline
H-DIBCO & DIBCO & Nabuco & Bickley-diary &Palm-Leaf& PNSR  \\ 
\hline
 & & &  &  & 14.26 \\ \hline
{\cmark} &  &  & & & 18.25\\ \hline
  & \cmark &  &  &   & 18.08\\ \hline
 \cmark &  \cmark&  &  &  & 18.10  \\ \hline
 \cmark  & \cmark &  \cmark&   &   & 16.89  \\ \hline
 \cmark & \cmark & &  \cmark&   &17.98   \\ \hline
 \cmark  & \cmark&  &    & \cmark& \textbf{21.85}    \\ \hline 
 & &  \cmark &  \cmark& \cmark&  14.78    \\ \hline
 \cmark  & \cmark& \cmark &  \cmark& \cmark& 17.63     \\ \hline

\end{tabular}
\end{table}
\FloatBarrier

\section{Conclusion}\label{section:conclusion}

In this paper, we proposed an architecture for handwritten document binarization  based on GANs. Our method recovers the degraded images while conserving its readability by integrating a HTR to evaluate the enhanced image in addition to the discriminator. To the best of our knowledge, this is the first approach that includes textual information when performing the recovery process of handwritten documents. Experimental results proved the effectiveness of the proposed model in cleaning extremely degraded documents.
We proved also that training a HTR model progressively on the images binarized by the  generator at each iteration leads to a better performance in CER and WER. We used in this paper a CRNN as a recognizer, but it is worth to note that by using other HTR architectures we may obtain a better recognition performances, since our method is flexible to integrate different ones. Moreover, we obtain the best performance compared to the state of the art in H-DIBCO  benchmarks of degraded documents binarization. 

Training supervised approaches using paired data (degraded image with its clean version and GT text) is costly, since it is really hard  to obtain this kind of resources. That is why most of the datasets are synthetically made. As a future work, we intend to explore the use of  unpaired data. In this way, the real images can be easily obtained (degraded  images, and clean ones). Thus,  our model could be trained  to produce images that are for  the discriminator as much real as possible (similar to  the clean real documents) and for  the recognizer as much readable and real as possible (the recognized text must have sense).


\section*{Acknowledgements}

This work has been partially supported by the Swedish Research Council (grant 2018-06074, DECRYPT), the Spanish project RTI2018-095645-B-C21, the Ramon y Cajal Fellowship RYC-2014-16831 and the CERCA Program / Generalitat de Catalunya.

\bibliography{main}

\end{document}